\documentclass[review]{elsarticle}
\usepackage{booktabs}
\usepackage{url}
\usepackage{epsfig}
\usepackage{subfigure}
\usepackage{graphics}
\usepackage{graphicx}
\usepackage{formular}
\usepackage{algorithm}
\usepackage{algorithmic}
\usepackage{multirow}
\usepackage{multicol}
\usepackage{xcolor}
\usepackage{clrscode}
\usepackage{makecell}
\usepackage{CJK}
\usepackage{amsthm}
\usepackage{balance}
\usepackage{times}
\usepackage{amsmath}
\usepackage{amsfonts}
\usepackage{stfloats}
\usepackage{array}
\usepackage{float}
\usepackage{url}
\usepackage{epstopdf}
\usepackage[colorlinks,linkcolor=blue]{hyperref}
\usepackage{hyperref}
\usepackage{lipsum}
\usepackage[figuresright]{rotating}
\usepackage{lineno,hyperref}
\usepackage{fontenc}
\modulolinenumbers[5]

\journal{Journal of Knowledge-Based Systems}

\begin{document}

\begin{frontmatter}

\title{Multi-local Collaborative AutoEncoder}

\author[mymainaddress,mysecondaryaddress]{Jielei Chu}


\author[mymainaddress,mysecondaryaddress]{Hongjun Wang\corref{mycorrespondingauthor}}
\cortext[mycorrespondingauthor]{Corresponding author at: School of Computing and Artificial Intelligence, Southwest Jiaotong University, China}
\ead{wanghongjun@swjtu.edu.cn}

\author[mythirdaryaddress]{Jing Liu}
\author[myfouraryaddress]{Zhiguo Gong}
\author[mymainaddress,mysecondaryaddress]{Tianrui Li}


\address[mymainaddress]{School of Computing and Artificial Intelligence, Southwest Jiaotong University, Chengdu 611756, China}
\address[mysecondaryaddress]{Institute of Artificial Intelligence, Southwest Jiaotong University, Chengdu 611756, China}
\address[mythirdaryaddress]{School of Business, Sichuan University, Sichuan, 610065, Chengdu, China}
\address[myfouraryaddress]{State Key Laboratory of Internet of Things for Smart City, Department of Computer and Information Science, University of Macau, Macau, China}

\begin{abstract}
The excellent performance of representation learning of autoencoders have attracted considerable interest in various applications. However, the structure and multi-local collaborative relationships of unlabeled data are ignored in their encoding procedure that limits the capability of feature extraction. This paper presents a Multi-local Collaborative AutoEncoder (MC-AE), which consists of novel multi-local collaborative representation RBM (mcrRBM) and multi-local collaborative representation GRBM (mcrGRBM) models. Here, the Locality Sensitive Hashing (LSH) method is used to divide the input data into multi-local cross blocks which contains multi-local collaborative relationships of the unlabeled data and features since the similar multi-local instances and features of the input data are divided into the same block. In mcrRBM and mcrGRBM models, the structure and multi-local collaborative relationships of unlabeled data are integrated into their encoding procedure. Then, the local hidden features converges on the center of each local collaborative block. Under the collaborative joint influence of each local block, the proposed MC-AE has powerful capability of representation learning for unsupervised clustering. However, our MC-AE model perhaps perform training process for a long time on the large-scale and high-dimensional datasets because more local collaborative blocks are integrate into it. Five most related deep models are compared with our MC-AE. The experimental results show that the proposed MC-AE has more excellent capabilities of collaborative representation and generalization than the contrastive deep models.
\end{abstract}


\end{frontmatter}


\section{Introduction}
 Autoencoders have shown promising capability of representation learning and attracted considerable interest in various applications (e.g., classifications \cite{9173645}, dictionary learning \cite{9144476} and clustering \cite{8742794}). Although autoencoders are capable of learning complex mappings, how to capture meaningful structure of the latent feature is a long-term challenge in deep learning. Various other deep learning methods have been successful applied in practical applications (e.g., multi-context socially-aware navigation \cite{DBLP}).\\
 \indent There are various excellent autoencoders have been proposed \cite{9173645, 9144476, 9314101}. Wang et al. \cite{9173645} presented a within-class scatter information constraint-based autoencoder (WSI-AE), which minimizes the within-class scatter and the reconstruction error. The WSI-AE reduces the meaningless encoded features of classical AEs and enhances the feature discriminability. For convolutional dictionary learning problems, Tolooshams et al. \cite{9144476} established a link between autoencoder and dictionary learning and proposed a constrained recurrent sparse autoencoder (CRsAE) model. For domain adaptation, Yang et al. \cite{9314101} developed a dual-representation autoencoder (DRAE), which has capability to learn dual-domain-invariant representations. The DRAE has three leaning phases: 1) learn global representation of all target and source data; 2) extract local representations of instances; 3) construct dual representations by aligning the local and global representations with different weights. For semi-supervised learning, $\acute{\texttt{S}}$mieja et al. \cite{9178469} presented a semi-supervised Gaussian Mixture Autoencoder (SeGMA), which has the capability to learn a joint probability distribution between the data and their classes. In the latent space, a mixture of Gaussians is chosen as a target distribution. To produce better data samples and use the class-bassed discriminating features, Karatsiolis and Schizas \cite{8924906} proposed a generative denoising autoencoder model, which is sampled with a Markov chain Monte Carlo process.\\
\indent More recently, some autoencoders based on generative adversarial model have been presented. Ge et al. \cite{8742794} developed a dual adversarial autoencoder (Dual-AAE) model, which simultaneously maximizes the mutual information and likelihood function to extract classification and structure information. To learn interpretable latent representations for undirected graphs, Kipf and Welling \cite{VGAE} developed the Variational Graph AutoEncoder (VGAE), a probabilistic generative model for unsupervised graph representation learning on graph-structured data. Because of its excellent representation learning capability, it is getting more and more attention for deep clustering \cite{XU2020106260} on the image data and classification \cite{8706960} on the medical data.\\
\indent Restricted Boltzmann Machines (RBMs) and relevant autoencoders have been proved to be provided with the capability of representation learning \cite{hinton2006reducing, hinton1986learning, hinton2002training, krizhevsky2009learning, 2019Improved, 2020Restricted, Chujielei2018pcGRBM, 9165942, dbnKnowledge, Hu2017A}. In our previous work \cite{Chujielei2018pcGRBM}, we also proposed a powerful variant of GRBM called pcGRBM for semi-supervised representation learning. The pairwise constraints are used to guiding the encoding procedure. In many practical applications of machine learning, the labels are scarce. Hence, we proposed a multi-clustering integration RBM (MIRBM) in \cite{9165942} and developed an unsupervised feature learning architecture with MIRBM. The experiments proved that our semi-supervised pcGRBM and unsupervised MIRBM have excellent capability of representation learning. However, the structure and collaborative relationships of unlabeled data have been ignored in their encoding procedure of the shallow models.\\
\indent In this paper, we focus on a novel Multi-local Collaborative AutoEncoder (MC-AE) to capture hidden features and learn collaborative representation. In the structure of the MC-AE, there are two novel variants of RBM: multi-local collaborative representation RBM (mcrRBM) and multi-local collaborative representation GRBM (mcrGRBM). First, the unlabeled input data of mcrRBM and mcrGRBM models are divided into multi-local cross blocks by the Locality Sensitive Hashing (LSH) \cite{Fran2016A} method in the dimensions of instances and features simultaneously. Then, the similar local instances and features of the input data are divided into the same block. Hence, these blocks contains multi-local collaborative relationships of the unlabeled data and feature. Furthermore, the local hidden features converge on the center of each local collaborative block in the encoding procedures of the mcrRBM and mcrGRBM models. Under the collaborative joint influence of each local block, the proposed MC-AE has powerful capability of representation learning for unsupervised clustering.\\
\indent This is the first work to capture hidden features and learn collaborative representation in autoencoder from the structure perspective of unlabeled data with multi-local collaborative relationship. The contributions can be summarized as follows:\\
\begin{itemize}
  \item One novel variant of RBM called multi-local collaborative representation RBM (mcrRBM) and another novel variant of GRBM called multi-local collaborative representation GRBM (mcrGRBM) are proposed by fusing the structure of unlabeled data with multi-local collaborative relationship to capture hidden features and learn collaborative representation in their encoding procedures.
  \item A novel Multi-local Collaborative AutoEncoder (MC-AE) based on mcrRBM and mcrGRBM are developed. For modeling real-valued data, one architecture of the MC-AE is composed of a mcrGRBM and two mcrRBMs which have Gaussian linear visible units and binary hidden layer units. For modeling binary data, another architecture of the MC-AE is composed of with three mcrRBMs which have binary visible and hidden units.
  \item The experiments demonstrate that the proposed MC-AE has powerful capability of collaborative representation than five contrastive models on real-valued and binary datasets. Furthermore, the hidden collaborative features of the MC-AE show generalization ability for different clustering algorithms.
\end{itemize}
\indent The remaining of the paper is organized as follows. The related works are introduced in Section \uppercase\expandafter{\romannumeral2}. The theoretical background is described in Section \uppercase\expandafter{\romannumeral3}. The Multi-local Collaborative AutoEncoder (MC-AE) is developed in Section \uppercase\expandafter{\romannumeral4}. The experimental framework is illustrated in Section \uppercase\expandafter{\romannumeral5}. The experimental results and discussions are shown in Section \uppercase\expandafter{\romannumeral6}. Finally, our contributions are summarized in Section \uppercase\expandafter{\romannumeral7}.
\section{Related Work}
Collaborative representation learning originates the influential sparse representation-based classification (SRC) \cite{4483511}. It has attracted much attentions in collaborative filtering \cite{8279660, 8674571}. \\
\indent Various deep networks based on classical RBM and variants are developed in practical applications \cite{cho2013gaussian, zhang2014supervised, 7346495} . There are some most relevant work: 1) deep autoencoder (DAE) \cite{hinton2006reducing}; 2) feature selection algorithm for Deep Boltzmann Machines (Deep-FS) \cite{TAHERKHANI201822}; 3) collaborative deep learning (CDL) \cite{WangWY15}; 4) full GraphRBM-based
DBN (fGraphDBN) \cite{7927417}. \\
\indent The DAE \cite{hinton2006reducing} as a classic unsupervised deep model consists of a stack of traditional RBMs for representation learning of binary data. It is also used to model real-valued data by replacing binary visible layer units with Gaussian linear visible units. Deep Boltzmann Machines (DBMs) \cite{srivastava2013modeling, salakhutdinov2009deep, salakhutdinov2012efficient} have reasonable structures to learn complex relationships between features. Taherkhani et al. \cite{TAHERKHANI201822} presented a Deep-FS model, which has powerful capability of removing irrelevant features from raw data to explore the underlying representations. Reducing irrelevant features is an important strategy to prevent negative impact in the encoding procedure. Under considering the local manifold structure of the data, Chen et al. \cite{7927417} developed a graph regularized RBM (GraphRBM) to learn hidden features. To obtain superior expressive power of deep model, an fGraphDBN model was developed using a stack of GraphRBM. However, none of them have collaborative representation capabilities. By adding a collaborative strategy, Wang et al. \cite{WangWY15} proposed a popular hierarchical Bayesian model, CDL, which jointly performs collaborative filtering and deep representation learning. In this paper, the structure and multi-local collaborative relationships of unlabeled data are fused into the encoding procedure of the proposed MC-AE. To prove the effectiveness of our models, we compare them with these most related works in the experiments.\\
\section{Theoretical Background}
\subsection{Restricted Boltzmann Machine}
 For classic RBMs \cite{hinton1986learning}, its architecture is a shallow two-layer structure, which consists of a binary visible and hidden layer. The RBM is an energy based model and the energy function of it is defined by:
   \begin{equation}
   E(\textbf{\texttt{v}},\textbf{\texttt{h}})=-\sum\limits_{i\in visibles}a_{i}v_{i}-\sum\limits_{j\in hiddens}b_{j}h_{j}-\sum\limits_{i,j}v_{i}h_{j}w_{ij},
   \end{equation}
  where $\textbf{\texttt{v}}$ and $\textbf{\texttt{h}}$ are the visible and hidden layer vectors, respectively, $v_{i}$ and $h_{j}$ are the binary visible and hidden units, respectively, $w_{ij}$ is the symmetric connection weight between them, $a_{i}$ and $b_{j}$ are the biases of visible and hidden units, respectively.\\
 \indent Given a visible vector $\textbf{\texttt{v}}$, the binary state $h_{j}$ is equal to 1 with probability
  \begin{equation}
   p(h_{j}=1|\texttt{\textbf{v}})=\sigma(b_{j}+\sum\limits_{i}v_{i}w_{ij}),
   \end{equation}
 where $\sigma(x)=\frac{1}{1+\texttt{exp}(-x))}$, which is a logistic sigmoid function.\\
 \indent Similarly, given a hidden vector $\textbf{\texttt{h}}$, an unbiased sample of the binary state $v_{i}$ is equal to 1 with probability
 \begin{equation}
   p(v_{i}=1|\texttt{\textbf{h}})=\sigma(a_{i}+\sum\limits_{j}h_{j}w_{ij}).
   \end{equation}
  It is difficult to get an unbiased sample of an average of the model distribution $<v_{i}h_{j}>_{model}$ because of low computing efficiency. Hinton proposed a faster learning algorithm by Contrastive
Divergence (CD) \cite{hinton2002training}, \cite{carreira2005contrastive} method. Then the update rules of parameters is given by:
\begin{equation}
   \Delta w_{i}{j}=\varepsilon(<v_{i}h_{j}>_{data}-<v_{i}h_{j}>_{recon}),
   \end{equation}
   \begin{equation}
   \Delta a_{i}=\varepsilon(<v_{i}>_{data}-<v_{i}>_{recon}),
   \end{equation}
   \begin{equation}
   \Delta b_{j}=\varepsilon(<h_{j}>_{data}-<h_{j}>_{recon}),
   \end{equation}
   where $\varepsilon$ is a learning rate, $<\cdot>_{data}$ is an average of the data distribution and $<\cdot>_{recon}$ is an average under the distribution of reconstructed units.
\subsection{Gaussian Linear Visible Units}
For modeling real-valued data, the binary visible units are replaced by Gaussian linear visible units. The energy function becomes:
\begin{equation}
\begin{aligned}
   E(\textbf{\texttt{v}},\textbf{\texttt{h}})=&\sum\limits_{i\in visibles}\frac{(v_{i}-a_{i})^{2}}{2\sigma_{i}^2}-\sum\limits_{j\in hiddens}b_{j}h_{j}\\
   &-\sum\limits_{i,j}\frac{v_{i}}{\sigma_{i}}h_{j}w_{ij},
   \end{aligned}
   \end{equation}
where $\sigma_{i}$ is the standard deviation of the Gaussian noise for visible unit $i$. It is difficult to use CD method to learn the variance of the noise. In practice, we normalise the original data to have unit variance and zero mean. So, the reconstructed result of a Gaussian linear visible unit is equal to the input from hidden binary units plus the bias.
\subsection{Locality Sensitive Hashing}
The Locality Sensitive Hashing (LSH) \cite{Fran2016A} exploits the probability that two similar samples likely collide by mapping with a weak hash function. In fact, the probability of the collision is proportional to their similarity. One classic hash function is the Minwise Independent Permutation (Minhash) \cite{v008a014} which defines the probability of collision is proportional to the Jaccard similarity of two hashed objects. The Jaccard similarity varies from 0 to 1. The value of it is 1 means that the two hashed objects are equal.
\section{Multi-local Collaborative AutoEncoder}
In this section, we firstly present the key basics of unsupervised Multi-local Collaborative AutoEncoder (MC-AE) that is the mcrRBM and mcrGRBM models (novel variants of RBM and GRBM). Then, we show the inference, learning algorithm and complexity analysis of the mcrRBM model. Finally, we propose two MC-AE deep architectures based on the mcrRBM and mcrGRBM models for modeling real-valued and binary data, respectively.
\subsection{The mcrRBM and mcrGRBM Models}
In this section, we present the key basics of the MC-AE deep architecture that is the mcrRBM (see Fig.1) and mcrGRBM models (see Fig.2). Here, we use the LSH \cite{Fran2016A} method to divide the input data into multi-local cross blocks with the perspective of instances and features simultaneously. Furthermore, the similar multi-local instances and features of the input data are divided into the same block. Then, we expect the hidden layer feature units of each block converges on the block center as much as possible in the encoding procedure of mcrRBM and mcrGRBM models. By this way, the correlations between the instances and features (multi-local collaborative relations) are fused in the hidden layer features. Due to the same mapping relations from the visible lay units to hidden layer units (sigmoid transformation) between the mcrRBM and mcrGRBM models, we only present the mcrRBM model and its inference and learning algorithm.\\
 \begin{figure}
  \centering
  \includegraphics[scale=0.445]{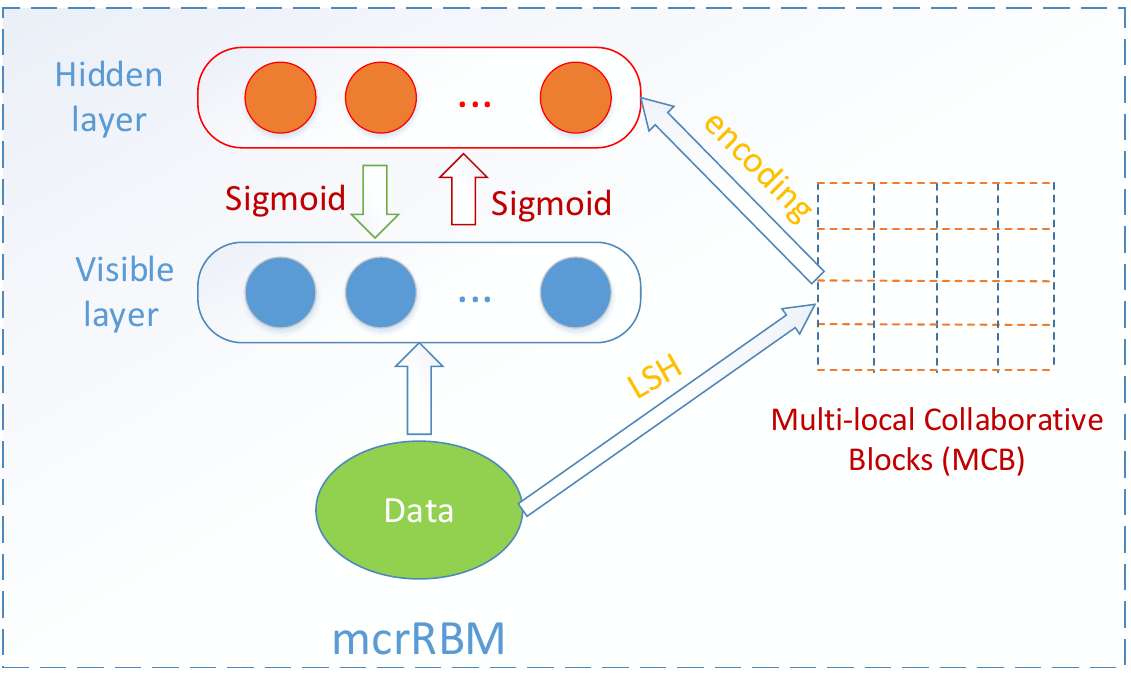}\\
  \caption{Multi-local collaborative representatin RBM (mcrRBM)}
\label{rbm}
\end{figure}
 \begin{figure}
  \centering
  \includegraphics[scale=0.445]{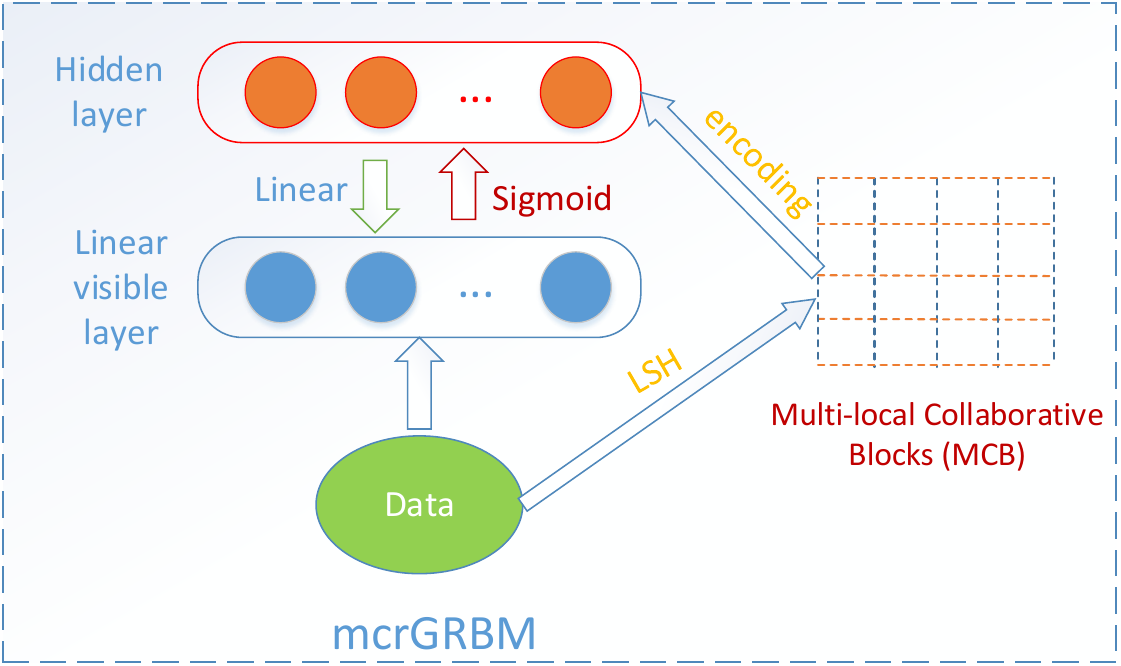}\\
  \caption{Multi-local collaborative representatin GRBM (mcrGRBM)}
\label{rbm}
\end{figure}
\indent Let $\mathcal D=\{\textbf{\texttt{\textbf{v}}}_{1},\textbf{\texttt{\textbf{v}}}_{2},\cdots,\textbf{\texttt{\textbf{v}}}_{N}\}$ be an original data set with $N$ vectors and $M$ features of each vector. The visible layer vector $\textbf{\texttt{\textbf{v}}}_{s}=(v_{s1},v_{s2},\cdots,v_{si},\cdots,v_{sM})$, $(i=1,2,\cdots,M$, and $s=1,2,\cdots,N)$. The hidden layer vector $\textbf{\texttt{h}}_{s}=(h_{s1}, h_{s2}, \cdots,h_{sj},\\
 \cdots, h_{sM'})$, $(j=1,2,\cdots,M'$, and $s=1,2,\cdots,N)$. The reconstructed visible layer vector $\textbf{\texttt{v}}_{s}^{(r)}=(v_{s1}^{(r)}, v_{s2}^{(r)},\cdots,v_{si}^{(r)},\cdots, v_{sM}^{(r)})$, $(i=1,2,\cdots,M$, and $s=1,2,\cdots,N)$. The reconstructed hidden layer vector $\textbf{\texttt{h}}_{s}^{(r)}=(h_{s1}^{(r)}, h_{s2}^{(r)},\cdots, h_{sj}^{(r)},\cdots,h_{sM}^{(r)})$, $(j=1,2,\cdots,M'$, and $s=1,2,\cdots,N)$. The matrix $(\textbf{\texttt{v}}_{1}^{T} \textbf{\texttt{v}}_{2}^{T} \cdots \textbf{\texttt{v}}_{N}^{T})^{T}$ is partitioned into $K$ row clusters by LSH and each cluster has a serial number set of vectors $\Re_{k}, (k=1,2,\cdots ,K$  and $\Re_{1} \cup \Re_{2}\cdots \cup  \Re_{K}=\{1,2,\cdots,N\})$. Simultaneously, the matrix is partitioned into $L$ column clusters by LSH and each cluster has a serial number set of vectors $\ell_{l}, (l=1,2,\cdots ,L$  and $ \ell_{1} \cup \ell_{2}\cdots \cup \ell_{K}=\{1,2,\cdots,M\})$. So, the matrix is divided into $K \times L$ blocks.\\
\indent Based on the expectations of our collaborative representation method and the training objective of classic RBM, our novel objective function takes the form:
\begin{equation}
\begin{aligned}
   &G(\texttt{V},\theta)=\\
   &-\frac{\eta}{N}\sum\limits_{\textbf{\texttt{v}}_{i}}\mathbf{log}^{p(\textbf{\texttt{v}}_{i};\theta)}+\frac{(1-\eta)}{K \times L}\Bigg[\sum\limits_{k=1}^{K}\sum\limits_{l=1}^{L}\sum\limits_{s\in \Re_{k}}\sum\limits_{t\in \ell_{l}}d(h_{st},u_{kl})\\
   &+\sum\limits_{k=1}^{K}\sum\limits_{l=1}^{L}\sum\limits_{s\in \Re_{k}}\sum\limits_{t\in \ell_{l}}d(h_{st}^{(r)},u_{kl}^{(r)})\Bigg],
\end{aligned}
\end{equation}
where $\theta=\{\mathbf{\textbf{a}},\mathbf{\textbf{b}},\mathbf{\textbf{\texttt{W}}}\}$ are the model parameters, $\eta$ is an adjusting parameter, $d(h_{st},u_{kl})$ and $d(h_{st}^{(r)},u_{kl}^{(r)})$ are the Bregman divergences \cite{Banerjee2005Clustering} distances, which are defined as: $d(h_{st},u_{kl})=(h_{st}-u_{kl})^2$ and $d(h_{st}^{(r)},u_{kl}^{(r)})=(h_{st}^{(r)}-u_{kl}^{(r)})^2$, respectively. $u_{kl}$ and $u_{kl}^{(r)}$ are the centers of block ($\Re_{k},\ell_{l}$) in hidden layer and reconstructed hidden layer, respectively. They take the form:
\begin{equation}
\begin{aligned}
   u_{kl}=\frac{\sum\limits_{s\in \Re_{k}}\sum\limits_{t\in \ell_{l}}h_{st}}{|\Re_{k}||\ell_{l}|}, u_{kl}^{(r)}=\frac{\sum\limits_{s\in \Re_{k}}\sum\limits_{t\in \ell_{l}} h_{st}^{(r)}}{|\Re_{k}||\ell_{l}|}.
\end{aligned}
\end{equation}
We expect all units close to their center of each local collaborative block in the representation learning process.
\subsection{The Inference}
In this subsection, we use the gradient descent algorithm to obtain the update rules of the parameters of the mcrRBM model. The detailed inference is shown as follows. \\
\indent Suppose that
\begin{equation}
\begin{aligned}
   C_{data}=\sum\limits_{k=1}^{K}\sum\limits_{l=1}^{L}\sum\limits_{s\in \Re_{k}}\sum\limits_{t\in \ell_{l}}(h_{st}-u_{kl})^2,\\
\end{aligned}
\end{equation}
\begin{equation}
\begin{aligned}
   C_{recon}=\sum\limits_{k=1}^{K}\sum\limits_{l=1}^{L}\sum\limits_{s\in \Re_{k}}\sum\limits_{t\in \ell_{l}}(h_{st}^{(r)}- u_{kl}^{(r)})^2.\\
\end{aligned}
\end{equation}
\indent Using the introduced variables $ C_{data}$ and $C_{recon}$, the objective function have another concise equivalent form:
 \begin{equation}
\begin{aligned}
   &G(\texttt{V},\theta)=\\
   &-\frac{\eta}{N}\sum\limits_{\textbf{v}_{i}}\mathbf{log}^{p(\textbf{v}_{i};\theta)}+\frac{(1-\eta)}{K \times L}(C_{data}+C_{recon}).
\end{aligned}
\end{equation}
\indent The following crucial problem is that how to solve this multi-objective optimization problem. For the average log-likelihood $\frac{\eta}{N}\sum\limits_{\textbf{v}_{i}}\mathbf{log}^{p(\textbf{v}_{i};\theta)}$, the CD method was presented to approximately follow the gradient of two divergences $\texttt{CD}_{n}=\texttt{KL}(p_{0}||p_{\infty})-\texttt{KL}(p_{n}||p_{\infty})$ to avoid enormous difficulties of the log-likelihood gradient computing. Normally, we run the Markov chain from the data distribution $p_{0}$ to $p_{1}$ (one step) in CD learning. So, the following key task is how to obtain the approximative gradient of $C_{data}+C_{recon}$. \\
\indent Suppose that $J_{data}=h_{st}- u_{kl}$,  $J_{recon}= h_{st}^{(r)}-  u_{kl}^{(r)}$, then
\begin{equation}
\begin{aligned}
   J_{data}&=h_{st}-\frac{\sum\limits_{s\in \Re_{k}}\sum\limits_{t\in \ell_{l}}h_{st}}{|\Re_{k}||\ell_{l}|}=\sigma\big(\sum\limits_{m=1}^{M}v_{sm}w_{mt}+b_{st}\big)\\
   &-\frac{\sum\limits_{s\in \Re_{k}}\sum\limits_{t\in \ell_{l}}\sigma\big(\sum\limits_{m=1}^{M}v_{sm}w_{mt}+b_{st}\big)}{|\Re_{k}||\ell_{l}|}
\end{aligned}
\end{equation}
and
\begin{equation}
\begin{aligned}
   J_{recon}&= h_{st}^{(r)}-\frac{\sum\limits_{s\in \Re_{k}}\sum\limits_{t\in \ell_{l}} h_{st}^{(r)}}{|\Re_{k}||\ell_{l}|}=\sigma\big(\sum\limits_{m=1}^{M} v_{sm}^{(r)}w_{mt}+ b_{st}^{(r)}\big)\\
   &-\frac{\sum\limits_{s\in \Re_{k}}\sum\limits_{t\in \ell_{l}}\sigma\big(\sum\limits_{m=1}^{M} v_{sm}^{(r)}w_{mt}+ b_{st}^{(r)}\big)}{|\Re_{k}||\ell_{l}|},
\end{aligned}
\end{equation}
where $\sigma$ is a sigmoid function.\\
When $t=j\in \ell_{l}$, the partial derivative of $J_{data}$ is given by:
\begin{equation}
\begin{aligned}
\frac {\partial J_{data}}{\partial w_{ij}}= &\frac{e^{-\big(\sum\limits_{m=1}^{M}v_{sm}w_{mj}+b_{sj}\big)}v_{si}}{\bigg[1+e^{-(\sum\limits_{m=1}^{M}v_{sm}w_{mj}+b_{sj})}\bigg]^{2}}\\
  &-\frac{\sum\limits_{s\in\Re_{k}}\frac{e^{-\big(\sum\limits_{m=1}^{M}v_{sm}w_{mj}+b_{sj}\big)}v_{si}}{\bigg[1+e^{-(\sum\limits_{m=1}^{M}v_{sm}w_{mj}+b_{sj})}\bigg]^{2}}}{|\Re_{k}|}\\
  =&(1-h_{sj})h_{sj}v_{si}-\frac{\sum\limits_{s\in \Re_{k}}(1-h_{sj})h_{sj}v_{si}}{|\Re_{k}|}
\end{aligned}
\end{equation}
Obviously, if $t\neq j$, then $\frac {\partial J_{data}}{\partial w_{ij}}=0$. \\
\indent Similarly, if $t=j\in \ell_{l}$, the partial derivative of $J_{recon}$ is given by:
\begin{equation}
\begin{aligned}
  &\frac {\partial J_{recon}}{\partial w_{ij}}=(1- h_{sj}^{(r)}) h_{sj}^{(r)} v_{si}^{(r)}-\frac{\sum\limits_{s\in \Re_{k}}(1- h_{sj}^{(r)}) h_{sj}^{(r)} v_{si}^{(r)}}{|\Re_{k}|}
 \end{aligned}
\end{equation}
As for model parameter $\textbf{\texttt{b}}$, if $t=j$, the partial derivative takes the forms:
\begin{equation}
\begin{aligned}
  \frac {\partial J_{data}}{\partial b_{j}}=(1-h_{sj})h_{sj}-\frac{\sum\limits_{s\in \Re_{k}}(1-h_{sj})h_{sj}}{|\Re_{k}|},\\
   \frac {\partial J_{recon}}{\partial b_{j}}=(1- h_{sj}^{(r)}) h_{sj}^{(r)}-\frac{\sum\limits_{s\in \Re_{k}}(1- h_{sj}^{(r)}) h_{sj}^{(r)}}{|\Re_{k}|}.
\end{aligned}
\end{equation}
\indent It is obvious that model parameter $\textbf{\texttt{a}}$ is independent of $J_{data}$ and $J_{recon}$. So, we can obtain that $\frac {\partial J_{data}}{\partial a_{i}}=0$ and  $\frac {\partial J_{recon}}{\partial a_{i}}=0$.
Then, the partial derivative of the $C_{data}$ in terms of $w_{ij}$ takes the form:
\begin{equation}
\begin{aligned}
 \frac {\partial C_{data}}{\partial w_{ij}}=&2\sum\limits_{k=1}^{K}\sum\limits_{s\in \Re_{k}}\big(h_{sj}-\frac{\sum\limits_{s\in \Re_{k}}h_{sj}}{|\Re_{k}|}\big)\Bigg[(1-h_{sj})h_{sj}v_{si}\\
  &-\frac{\sum\limits_{s\in \Re_{k}}(1-h_{sj})h_{sj}v_{si}}{|\Re_{k}|}\Bigg].
\end{aligned}
\end{equation}
And the partial derivative of the $C_{recon}$ in terms of $w_{ij}$ takes the form:
\begin{equation}
\begin{aligned}
  \frac {\partial C_{recon}}{\partial w_{ij}}=&2\sum\limits_{k=1}^{K}\sum\limits_{s\in \Re_{k}}\big( h_{sj}^{(r)}-\frac{\sum\limits_{s\in \Re_{k}}h_{sj}^{(r)}}{|\Re_{k}|}\big)\Bigg[(1- h_{sj}^{(r)}) h_{sj}^{(r)} v_{si}^{(r)}\\
  &-\frac{\sum\limits_{s\in \Re_{k}}(1- h_{sj}^{(r)}) h_{sj}^{(r)} v_{si}^{(r)}}{|\Re_{k}|}\Bigg].
\end{aligned}
\end{equation}
Similarly, the partial derivative of the $C_{data}$ in terms of $b_{j}$ is given by:
\begin{equation}
\begin{aligned}
  \frac {\partial C_{data}}{\partial b_{j}}=&2\sum\limits_{k=1}^{K}\sum\limits_{s\in \Re_{k}}\big(h_{sj}-\frac{\sum\limits_{s\in \Re_{k}}h_{sj}}{|\Re_{k}|}\big)\Bigg[(1-h_{sj})h_{sj}\\
  &-\frac{\sum\limits_{s\in \Re_{k}}(1-h_{sj})h_{sj}}{|\Re_{k}|}\Bigg].
\end{aligned}
\end{equation}
And the partial derivative of the $C_{recon}$ in terms of $b_{j}$ is given by:
\begin{equation}
\begin{aligned}
  \frac {\partial C_{recon}}{\partial b_{j}}=&2\sum\limits_{k=1}^{K}\sum\limits_{s\in \Re_{k}}\big( h_{sj}^{(r)}-\frac{\sum\limits_{s\in \Re_{k}}h_{sj}^{(r)}}{|\Re_{k}|}\big)\Bigg[(1- h_{sj}^{(r)}) h_{sj}^{(r)}\\
  &-\frac{\sum\limits_{s\in \Re_{k}}(1-h_{sj}^{(r)}) h_{sj}^{(r)}}{|\Re_{k}|}\Bigg].
\end{aligned}
\end{equation}
\indent Combined with the CD learning with 1 step Gibbs sampling, the update rule of the proposed model parameter $\texttt{\textbf{W}}$ takes the forms:
\begin{equation}
\begin{aligned}
   &w_{ij}^{(\tau+1)}=w_{ij}^{(\tau)}+\eta\varepsilon(<v_{i}h_{j}>_{data}-<v_{i}h_{j}>_{recon})\\
   &+\frac{2(1-\eta)}{K \times L}\Bigg\{\sum\limits_{k=1}^{K}\sum\limits_{s\in \Re_{k}}\big(h_{sj}-\frac{\sum\limits_{s\in \Re_{k}}h_{sj}}{|\Re_{k}|}\big)\Bigg[(1-h_{sj})h_{sj}v_{si}\\
   &-\frac{\sum\limits_{s\in \Re_{k}}(1-h_{sj})h_{sj}v_{si}}{|\Re_{k}|}\Bigg]+\sum\limits_{k=1}^{K}\sum\limits_{s\in \Re_{k}}\big(h_{sj}^{(r)}-\frac{\sum\limits_{s\in \Re_{k}}h_{sj}^{(r)}}{|\Re_{k}|}\big)\\
   &\Bigg[(1- h_{sj}^{(r)}) h_{sj}^{(r)} v_{si}^{(r)}-\frac{\sum\limits_{s\in \Re_{k}}(1-h_{sj}^{(r)}) h_{sj}^{(r)} v_{si}^{(r)}}{|\Re_{k}|}\Bigg]\Bigg\},
   \end{aligned}
\end{equation}
where $\varepsilon$ is learning rate, the average $<v_{i}h_{j}>_{data}$ and $<v_{i}h_{j}>_{recon}$ are computed using the sample data and reconstructed data, respectively.\\
\indent For the parameters of the biases $\texttt{\textbf{a}}$ and $\texttt{\textbf{b}}$, the update rules of them take the forms:
\begin{equation}
\begin{aligned}
 a_{i}^{(\tau+1)}=a_{i}^{(\tau)}+\eta\varepsilon(<v_{i}>_{data}-<v_{i}>_{recon}),
   \end{aligned}
\end{equation}
and
\begin{equation}
\begin{aligned}
  &b_{j}^{(\tau+1)}=b_{j}^{(\tau)}+\eta\varepsilon(<h_{j}>_{data}-<h_{j}>_{recon})\\
    &+\frac{2(1-\eta)}{K \times L}\Bigg\{\sum\limits_{k=1}^{K}\sum\limits_{s\in \Re_{k}}\big(h_{sj}-\frac{\sum\limits_{s\in \Re_{k}}h_{sj}}{|\Re_{k}|}\big)\Bigg[(1-h_{sj})h_{sj}\\
  &-\frac{\sum\limits_{s\in \Re_{k}}(1-h_{sj})h_{sj}}{|\Re_{k}|}\Bigg] +
  \sum\limits_{k=1}^{K}\sum\limits_{s\in \Re_{k}}\big( h_{sj}^{(r)}-\frac{\sum\limits_{s\in \Re_{k}}h_{sj}^{(r)}}{|\Re_{k}|}\big)\\
  &\Bigg[(1-h_{sj}^{(r)}) h_{sj}^{(r)}-\frac{\sum\limits_{s\in \Re_{k}}(1-h_{sj}^{(r)}) h_{sj}^{(r)}}{|\Re_{k}|}\Bigg]\Bigg\}.\\
   \end{aligned}
\end{equation}
\subsection{The Algorithm}
\textbf{Algorithm 1 Learning algorithm of mcrRBM with 1 step Gibbs sampling}\\
\noindent\line(1,0){250}\\
\textbf{Input}: $\mathcal D$: input data sets;\\
\indent \indent $\mathcal B$: training batch sets;\\
\indent \indent  $\varepsilon$: learning rate;\\
\indent \indent ($\Re_{k},\ell_{l}$): matrix blocks of $\mathcal D$, $k\in[1,K]$ and $l\in[1,L]$;\\
\noindent\line(1,0){250}\\
\textbf{Output}: $\theta$: model parameters of mcrRBM.\\
\noindent\line(1,0){250}\\
\indent Initialize: $\mathbf{a}$, $\mathbf{b}$ and $\mathbf{W}$.\\
\indent \textbf{while} $\tau$ not exceeding maximum iteration {do}\\
\indent \indent \textbf{for} all training batch $\mathcal B$ {do}\\
\indent \indent \indent Encoder: sample the states of hidden units by  \\
\indent \indent \indent \indent \indent $p(h_{j}=1|\texttt{\textbf{v}})=\sigma(b_{j}+\sum\limits_{i}v_{i}w_{ij})$.\\
\indent \indent \indent Decoder: sample the reconstructed states of visible units using\\
\indent \indent \indent \indent \indent  $p(v_{i}=1|\texttt{\textbf{h}})=\sigma(a_{i}+\sum\limits_{j}h_{j}w_{ij})$.\\
\indent \indent \indent \textbf{for} all $\Re_{k}$ {do}\\
\indent \indent \indent \indent \textbf{for} all $\ell_{l}$ {do}\\
\indent \indent \indent \indent \indent Compute the partial derivative $\frac {\partial C_{data}}{\partial w_{ij}}$ using Eq. (18).\\
\indent \indent \indent \indent \indent Compute the partial derivative $\frac {\partial C_{recon}}{\partial w_{ij}}$ using Eq. (19).\\
\indent \indent \indent \indent \indent Compute the partial derivative $\frac {\partial C_{data}}{\partial b_{j}}$ using Eq. (20).\\
\indent \indent \indent \indent \indent Compute the partial derivative $\frac {\partial C_{recon}}{\partial b_{j}}$ using Eq. (21).\\
\indent \indent \indent \indent \textbf{end for}\\
\indent \indent \indent \textbf{end for}\\
\indent \indent \indent Update parameter $\mathbf{W}$ using Eq. (22).\\
\indent \indent \indent Update parameter $\mathbf{a}$ using Eq. (23).\\
\indent \indent \indent Update parameter $\mathbf{b}$ using Eq. (24).\\
\indent \indent \textbf{end for}\\
\indent \indent $\tau=\tau+1$.\\
\indent \textbf{end while}\\
\indent  \textbf{return} $\mathbf{a}$, $\mathbf{b}$ and $\mathbf{W}$.\\
\noindent\line(1,0){250}\\
 \indent In the reconstruction process of mcrGRBM model, a linear reconstruction method replaces the nolinear reconstruction method of mcrRBM model. The steps of the learning algorithms of our mcrRBM and mcrGRBM models are almost the same, except the reconstruction process. So, we omit the learning algorithm of mcrGRBM model.
\subsection{Complexity Analysis}
In this subsection, we analyze the time complexity of above learning algorithm. Supposing that the input data sets $\mathcal D$ is divided into $TB$ training batch. Then the time complexities of the encoder and decoder steps are $O(TB)$ in each iteration. When partial derivatives $\frac {\partial C_{data}}{\partial w_{ij}}$, $\frac {\partial C_{recon}}{\partial w_{ij}}$, $\frac {\partial C_{data}}{\partial b_{j}}$ and $\frac {\partial C_{recon}}{\partial b_{j}}$ are calculated, they take $O(TB\times (K \times L))$ in each iteration. The complexities of update parameters $\mathbf{W}$, $\mathbf{a}$ and $\mathbf{b}$ are $O(TB)$ in each iteration. Supposing that the maximum iteration is $IT$. Then, the time complexity of the mcrRBM learning algorithm with 1 Step Gibbs sampling is $O(IT\times TB\times (K \times L))$.
\subsection{The MC-AE Deep Architecture}
A novel Multi-local Collaborative AutoEncoder (MC-AE) architecture is developed with one visible layer and three hidden layer (see Fig. 3). To learn collaborative representation of two types input data (binary and real-valued), the visible layer units can be designed as binary and linear units, respectively. In other words, one architecture of MC-AE for modeling binary data consists of three mcrRBM. And another architecture of MC-AE for modeling real-valued data consists of one mcrGRBM and two mcrRBMs. In the encoding procedure, the first multi-local collaborative blocks (MCB 1) comes from raw data by LSH method. Then, local collaborative relationships of the unlabeled data and feature force the local hidden features to converge on the center of each local collaborative block. The second multi-local collaborative blocks (MCB 2) is generated by LSH method from the first hidden layer. Similarly, they are fused into the second hidden layer, and so forth. The next experiments confirm the collaborative joint influence of each local block to improve the capability of representation learning of the proposed MC-AE.
 \begin{figure}[h]
  \centering
  \includegraphics[scale=0.345]{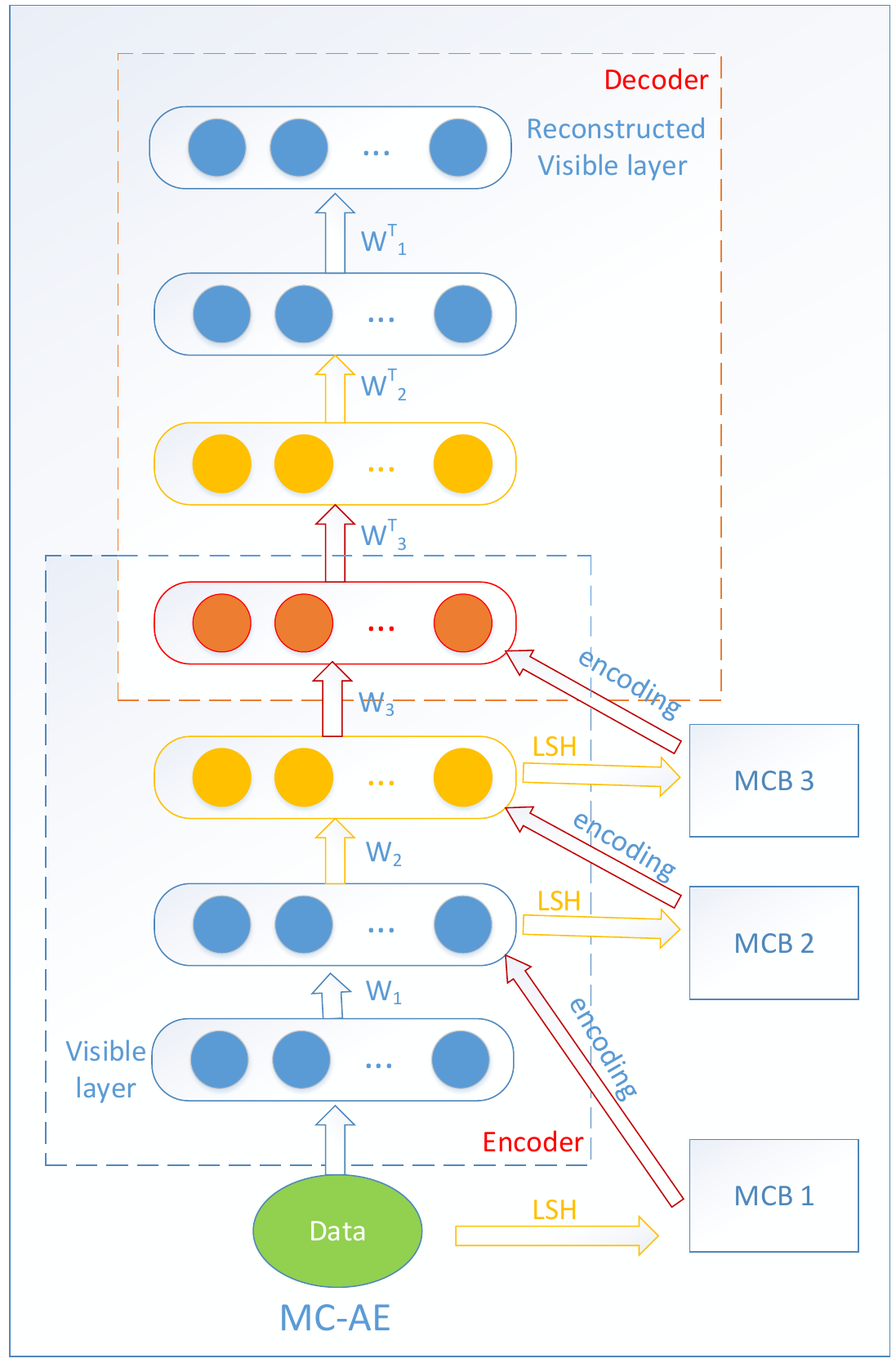}\\
  \caption{Multi-local Collaborative AutoEncoder (MC-AE). One architecture of MC-AE (Linear visible layer and binary hidden layer) consists of one mcrGRBM and two mcrRBMs for modeling real-valued data. Another architecture of MC-AE (visible and hidden layer units are both binary) consists of three mcrRBMs for modeling binary data. }
\label{rbm}
\end{figure}
\section{Experimental Framework}
\indent This section introduces the experimental datasets used in the current work, the experimental settings and the evaluation metrics.
\subsection{Datasets}
To explore the collaborative representation capability of the proposed MC-AE for real-valued data, we do experiments on ten image datasets from MSRA-MM 2.0 \cite{li2009msra}. The summaries of them are listed from No. 1 to No. 10 in Table 1. All of them have same class, but different instances and features. The datasets of banner, beret, bugatti and building have 892 features, but the vista, vistawallpaper, water, wing and worldmap have 899 features. To explore the collaborative representation capability of our MC-AE for binary data further, we do experiments on ten UCI datasets \footnote[1]{http://archive.ics.uci.edu/ml/index.php}. The summaries of them are listed from No. 11 to No. 20 in Table 1. They have different classes, instances and features.
\begin{table}[h]
\begin{center}
\caption{Experimental datasets}
 \label{imagedata}
\scalebox{0.63}{
\begin{tabular}{lcccc}
\toprule[1pt] 
{No.} &\textsf{\bf{Dataset}} &  {classes}&  {Instances} &  {features} \\
\toprule[1.5pt]
{1} & \textsf{banner} & {3} & {860}  & {892} \\
{2} &\textsf{beret} & {3} & {876} &{892} \\
{3} &\textsf{bugatti} & {3} & {882}  & {892} \\
{4} &\textsf{building}& {3} & {911} &{892}\\
{5} &\textsf{vista} & {3} & {799} &{899} \\
{6} &\textsf{vistawallpaper} & {3} & {799} &{899} \\
{7} &\textsf{voituretuning} & {3} & {879} &{899} \\
{8} &\textsf{water} & {3} & {922} &{899} \\
{9} &\textsf{wing} & {3} & {856} &{899} \\
{10} &\textsf{worldmap} & {3} & {935} &{899} \\
\hline
{11} & \textsf{balance} & {3} & {625}  & {4} \\
{12} & \textsf{biodegradation} & {2} & {1055}  & {41} \\
{13} &\textsf{car} & {4} & {1728} &{6} \\
{14} & \textsf{Climate Model} & {2} & {540} &{18} \\
{15} &\textsf{dermatology} & {6} & {366}  & {34} \\
{16} &\textsf{Haberman Survival} & {2} & {306} &{3} \\
{17} &\textsf{Kdd (1999 partial data )} & {3} & {1280} &{41} \\
{18} &\textsf{Ozone Level Detection} & {2} & {2534} &{72} \\
{19} &\textsf{parkinsons} & {2} & {195} &{22} \\
{20} &\textsf{secom} & {2} & {1567} &{590} \\
\bottomrule[1pt] 
\end{tabular}}
\end{center} 	
\end{table}
\subsection{Experimental Settings}
\indent To validate the capability of collaborative representation of the proposed MC-AE, we compare it with the DAE \cite{hinton2006reducing}, Deep-FS \cite{TAHERKHANI201822}, fGraphDBN \cite{7927417} and VGAE \cite{VGAE}, which have not collaborative representation strategy. Furthermore, we compare our MC-AE with CDL \cite{WangWY15} model, which has collaborative representation strategy. For modeling real-valued data, the MC-AE architecture consists of three binary hidden layer and one linear visible layer in Fig. 3. The transformation functions of all hidden layers between encoding and decoding are sigmoid functions. But, the transformation functions of encoding and decoding between visible layer and the first hidden layer are sigmoid and linear functions, respectively. For modeling binary data, the MC-AE architecture consists of three binary hidden layers and one binary visible layer in Fig. 3. In other words, all transformation functions are both sigmoid function.\\
\indent To compare the generalization capabilities of our MC-AE for representation learning, two different unsupervised clustering algorithms: K-means \cite{Lloyd1982Least} and Spectral Clustering (SC) \cite{ng2002spectral} are applied to clustering task with the representation of the deepest hidden layers of all contrastive deep models. The clustering algorithms based on the DAE, Deep-FS, fGraphDBN, CDL, VGAE and our MC-AE models using K-means are called DAE+KM, Deep-FS+KM, fGraphDBN+KM, CDL+KM, VGAE+KM and MC-AE+KM, respectively. Similarly, the clustering algorithms based on the DAE, Deep-FS, fGraphDBN, CDL, VGAE and the proposed MC-AE models using SC are called DAE+SC, Deep-FS+SC, fGraphDBN+SC, CDL+SC, VGAE+SC and MC-AE+SC, respectively.\\
\indent In two frameworks of the proposed MC-AE, the dimensionality of each hidden and visible layer is same as the raw data. The learning rate and $\eta$ of our MC-AE are set to 0.001 and 0.1, respectively. The parameters of other contrastive models adopt the values suggested in their papers.\\
\indent Our MC-AE+KM and MC-AE+SC methods are implemented in Matlab 2016 (a). All contrastive methods have run on a Server with Core i9 CPU and 64 GB RAM.
\subsection{Evaluation Metrics}
In this paper, three classical clustering evaluation metrics: clustering accuracy (ACC) \cite{Cai2005Document}, Jaccard index (Jac) \cite{Fran2016A} and Fowlkes-Mallows index (FMI) \cite{LIU2018200} are utilized to evaluate the performance of the proposed MC-AE model. Furthermore, the Friedman Aligned Ranks test statistic \cite{garcia2010advanced} is used to report significant differences of all contrastive algorithms. The ACC evaluation metric takes the form:
\begin{equation}
\begin{aligned}
    accuracy=\frac{\sum\limits_{i=1}\delta(s_{i},map(r_{i}))}{n},
\end{aligned}
\end{equation}
where $map(r_{i})$ maps label $r_{i}$ of each cluster to the equivalent label and $n$ is the total number of instances. If $x=y$ , then $\delta(x,y)$ equals to 1 . Otherwise, it is zero. The Jac evaluation metric is defined by:
\begin{equation}
\begin{aligned}
    Jac=\frac{|A\cap B|}{|A\cup B|},
\end{aligned}
\end{equation}
where $A$ and $B$ are finite sample sets and $ 0\leq J(A,B)\leq 1$. The FMI evaluation metric is given by:
\begin{equation}
\begin{aligned}
  FMI=\sqrt{\frac{TP}{TP+FP}\times\frac{TP}{TP+FN}},
   \end{aligned}
\end{equation}
where $TP$ is the number of true positives, $FP$ is the number of false positives and and $FN$ is the number of false negatives.\\
\indent The Friedman Aligned Ranks test statistic \cite{garcia2010advanced} takes the form:
 \begin{equation}
\begin{aligned}
  T=\frac{(n_{a}-1)(\sum\limits_{j=1}^{n_{a}}\widehat{r}_{.j}^2-n_{a}n_{d}^2(n_{a}n_{d}+1)^2/4)}{n_{a}n_{d}(n_{a}n_{d}+1)(2n_{a}n_{d}+1)/6-\sum\limits_{i=1}^{n_{d}}\widehat{r}_{i.}^2/n_{a}},
\end{aligned}
\end{equation}
 where $\widehat{r}_{i.}$ and $\widehat{r}_{.j}$ are the ranks total of the $j$th algorithm and $i$th data set, respectively, $n_{a}$ and $n_{d}$ are the numbers of algorithm and data set, respectively. For $n_{a}-1$ degrees of freedom, the test statistic $T$ is compared for significance with a chi-square distribution.\\
\section{Reults and Discussion}
For fairness of comparisons, our MC-AE+KM algorithms based on the proposed MC-AE model are compared with DAE+KM, Deep-FS+KM, fGraphDBN+KM, CDL+KM and VGAE+KM, respectively. Similarly, our MC-AE+SC algorithms based on the proposed MC-AE model are compared with DAE+SC, Deep-FS+SC, fGraphDBN+SC, CDL+SC and VGAE+SC, respectively. Moreover, the results (ACC, Jac and FMI) of K-means and SC algorithms on original real-valued datasets and UCI datasets are listed in Table 10 and Table 11 for comparisons, respectively.
\subsection{Representation Learning for clustering on Real-valued Datasets}
\subsubsection{Accuracy}
 Table 2 shows the results of the ACC (mean$\pm$std) of each contrastive algorithm on each dataset and the average ACC ($\overline{ACC}$) of each algorithm is listed in the last column. The MC-AE+KM algorithm based on the proposed MC-AE shows the best performance on the banner, beret, building, vista, voituretuning and wing datasets. The ACC of them are 0.9372, 0.6895, 0.7164, 0.6308, 0.6394 and 0.6192, respectively. The MC-AE+SC algorithm based on the proposed MC-AE shows the best performance on the bugatti, vistawallpaper, water and woldmap datasets. The ACC of them are 0.7007, 0.6320, 0.5705 and 0.7134, respectively. The average ACC of MC-AE+KM and MC-AE+SC algorithms are 0.6335 and 0.6500, respectively. They show the best performance in the corresponding comparative grouping.  \\
\indent An intuitive comparison of the overall performance (average ACC) is shown in Fig. 5 (left one). From Table 2 and Fig. 5, we can draw the conclusion that the proposed MC-AE shows the better performance than other deep models (DAE, Deep-FS, fGraphDBN, CDL and VGAE) in terms of the ACC metric.
\subsubsection{Jaccard Index}
Table 4 shows the results of the Jac of each contrastive algorithm on each dataset and the average Jac ($\overline{Jac}$) of each algorithm is listed in the last column. The MC-AE+KM algorithm based on the proposed MC-AE shows the best performance on the banner, beret, building, vista, voituretuning and wing datasets. The Jac of them are 0.8820, 0.5348, 0.5574, 0.4738, 0.4760 and 0.4714, respectively. The MC-AE+SC algorithm based on the proposed MC-AE shows the best performance on the bugatti, vistawallpaper, water and woldmap datasets. The Jac of them are 0.5420, 0.4736, 0.4351 and 0.5602, respectively. The average Jac of MC-AE+KM and MC-AE+SC algorithms are 0.4771 and 0.5311, respectively. They also show most excellent performance in the corresponding comparative grouping.  \\
\indent The intuitive comparison of overall performance (average Jac) is shown in Fig. 5 (middle one). Therefore, we can draw the conclusion that the proposed MC-AE shows the best performance among all contrastive deep models in terms of the Jac metric from Table 4 and Fig.5.
\subsubsection{Fowlkes and Mallows Index}
Table 5 shows the results of the FMI of each contrastive algorithm on each dataset and the average FMI ($\overline{FMI}$) of each algorithm is listed in the last column. The MC-AE+KM algorithm based on the proposed MC-AE shows the best performance on the banner, beret, building, vista, voituretuning and wing datasets. The FMI of them are 0.9392, 0.7313, 0.7466, 0.6870, 0.6899 and 0.6866, respectively. The MC-AE+SC algorithm based on the proposed MC-AE shows the best performance on the bugatti, vistawallpaper, water and woldmap datasets. The FMI of them are 0.7279, 0.6865, 0.6585 and 0.7460, respectively. The average FMI of MC-AE+KM and MC-AE+SC algorithms are 0.6542 and 0.7172, respectively. They show the best performance in the corresponding comparative grouping.  \\
\indent An intuitive comparison of average FMI is shown in Fig. 5 (right one). In terms of the FMI metric, we also can draw the conclusion that the proposed MC-AE shows the better performance than DAE, Deep-FS, fGraphDBN, CDL and VGAE deep models from Fig. 5 and Table 5.
\subsubsection{The Friedman Aligned Ranks Test Statistic}
Table 3 shows the ranks (in the parentheses) and average ranks of all contrastive algorithms. The smaller rank means the better performance of the algorithm on the corresponding dataset. The average ranks of MC-AE+SC and MC-AE+KM algorithms based on our MC-AE are 15.7 and 16.8, respectively. However, the average ranks of DAE+KM, Deep-FS+KM, fGraphDBN+KM, CDL+KM and VGAE+KM are 95.2, 74.9, 71.45, 77.25 and 76.7, respectively. And the ranks of DAE+SC, Deep-FS+SC, fGraphDBN+SC, CDL+SC and VGAE+SC are 110.9, 53.9, 65.9, 37.8 and 29.5, respectively. Clearly, MC-AE+SC and MC-AE+KM algorithms show the best performance in the corresponding comparative grouping. By means of the Friedman Aligned test statistic, \emph{T}=7.7217 is the chi-square distribution with 11 degrees of freedom. The \emph{p}-value is $0.00000457$ which is computed by $\chi^2(11)$ distribution for one tailed test and the two-tailed probability is 0.00000913. Then, the null hypothesis is rejected at a high level significance. The \emph{p}-values are far less than 0.05, so the experimental results of algorithms are different.\\
 \subsubsection{Friedman + Post-hoc Nemenyi Tests}
 The results of Friedman test + post-hoc Nemenyi test \cite{2021A} are shown in Fig. 4 among all contrastive methods on real-valued datasets. It is obvious that the test values of MC-AE+KM based on our MC-AE model versus DAE+KM, Deep-FS+KM, fGraphDBN+KM, CDL+KM and VGAE+KM are less than significance level (5\%). Hence, there are striking differences between MC-AE+KM and five related contrastive methods (DAE+KM, Deep-FS+KM, fGraphDBN+KM, CDL+KM and VGAE+KM). In Fig. 4, most of the test values between our MC-AE+SC and five related contrastive methods (DAE+SC Deep-FS+SC, fGraphDBN+SC, CDL+SC and VGAE+SC) are less than 5\% significance level expect for the results of MC-AE+SC versus CDL+SC and VGAE+SC methods. So, although the MC-AE+SC method based on MC-AE model has better performance than CDL+SC and VGAE+SC methods, there are no significant difference between MC-AE+SC and them.
\begin{figure*}
\vspace{0.5mm} \centering
    \includegraphics[scale=0.6025]{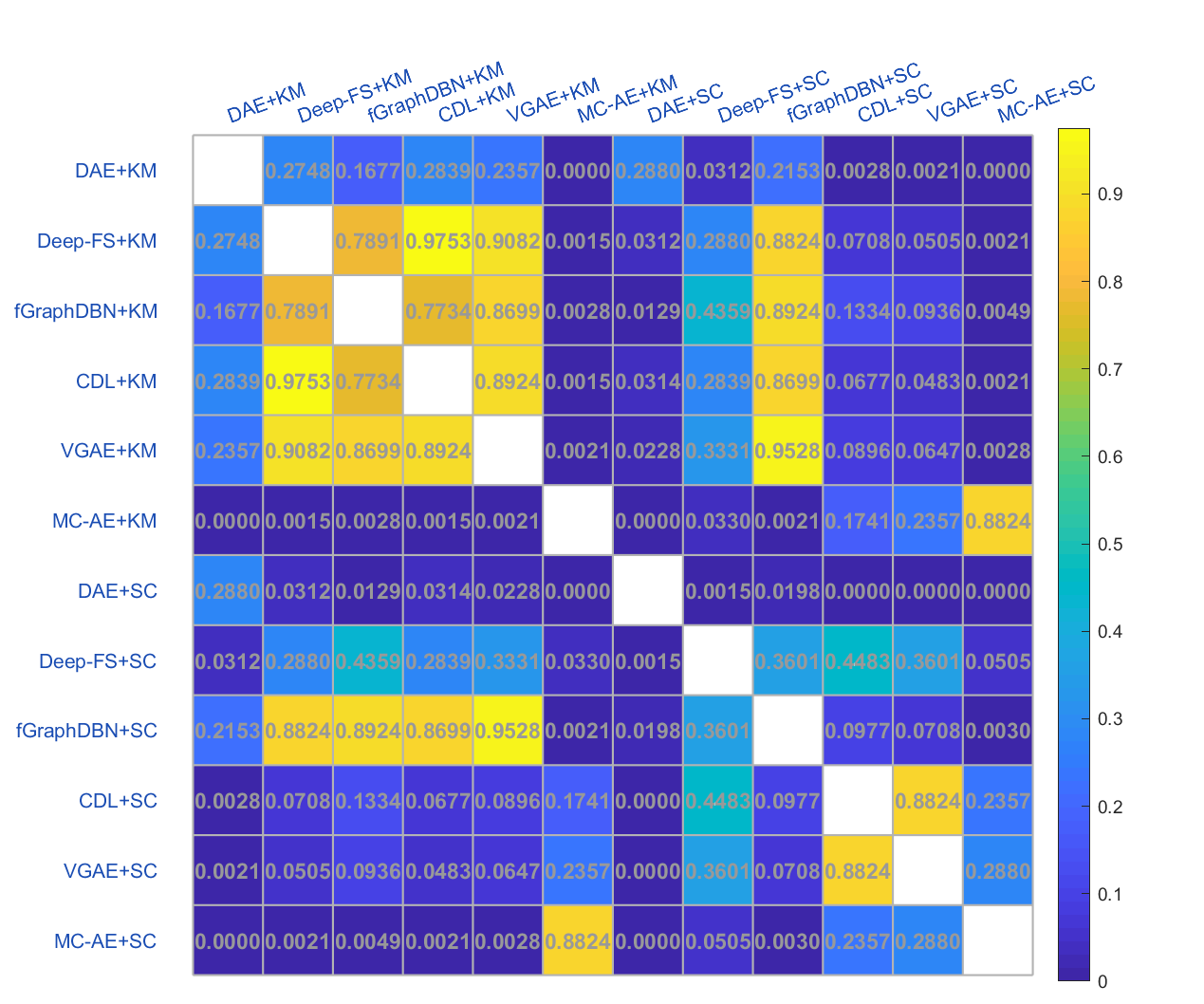}
\\
\caption{The results of Friedman + post-hoc Nemenyi tests among all contrastive algorithms on real-valued datasets.
} \label{fig:1}
\end{figure*}
\begin{figure*}
\vspace{0.5mm} \centering
    \includegraphics[scale=0.269025]{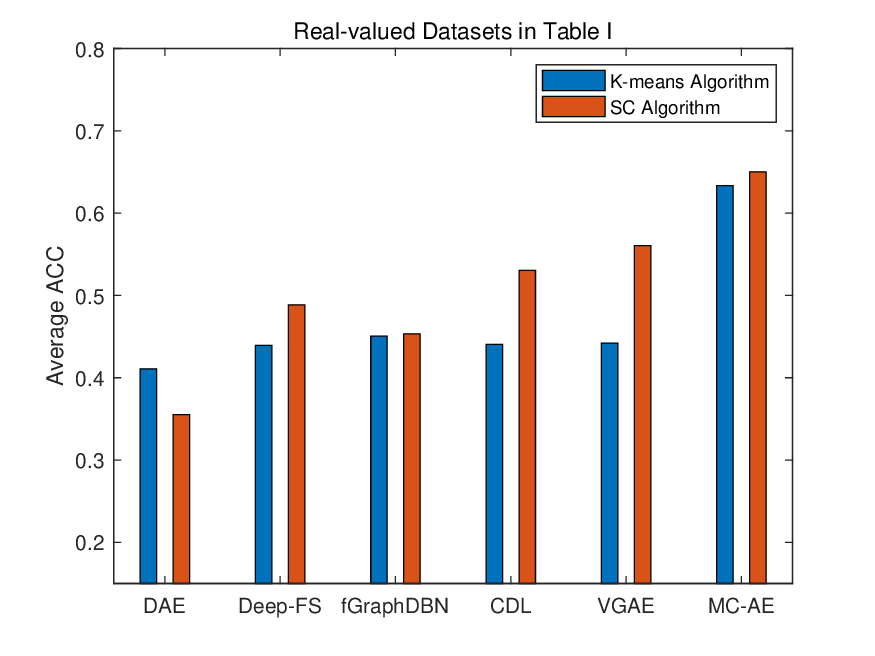}
    \includegraphics[scale=0.269025]{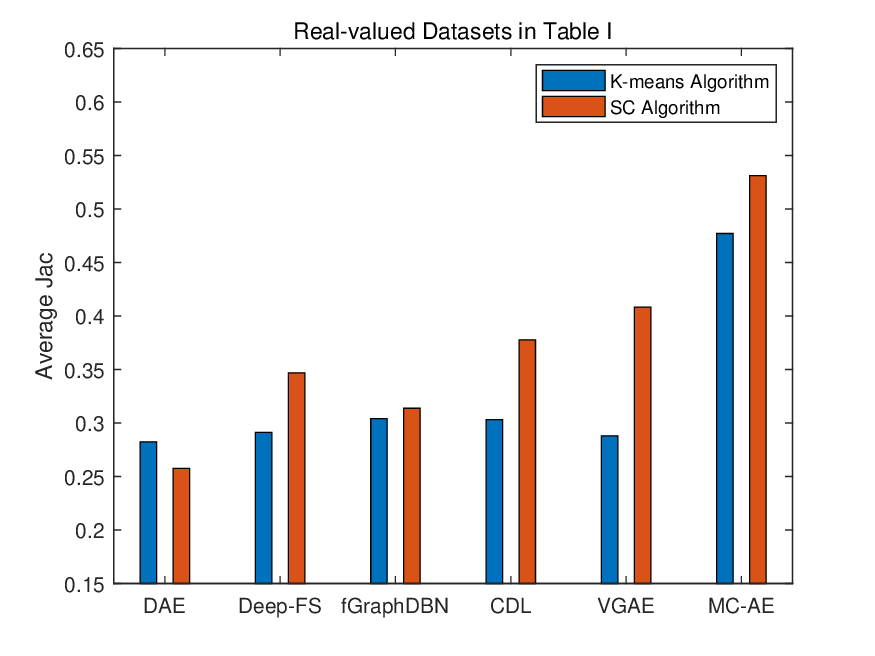}
    \includegraphics[scale=0.269025]{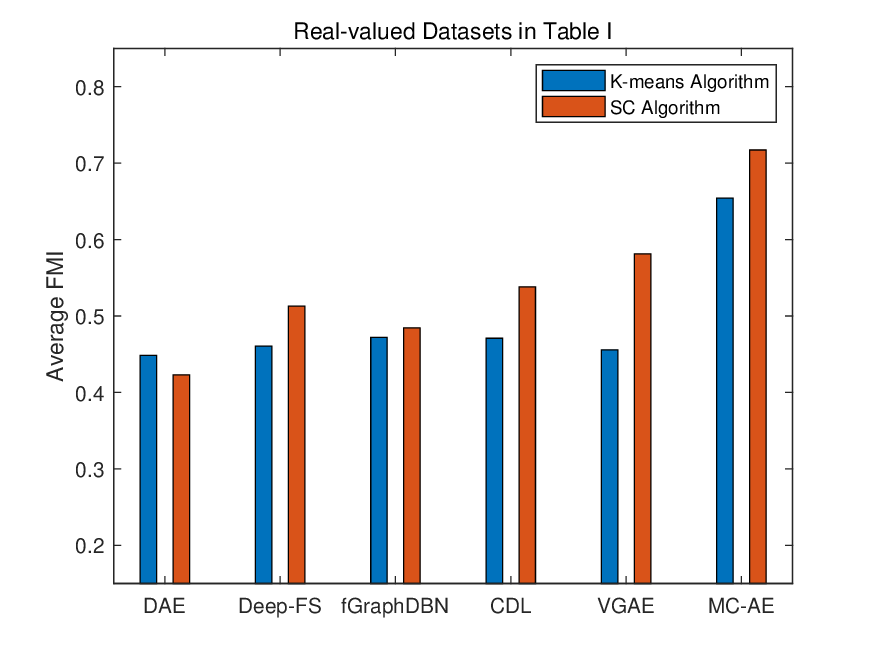}
\\
\caption{The performance comparisons of DAE, Deep-FS, fGraphDBN, CDL, VGAE and our MC-AE using average ACC, Jac and FMI metrics on the real-valued datasets.
} \label{fig:1}
\end{figure*}
\subsection{Representation Learning for Clustering on UCI Datasets}
\subsubsection{Accuracy}
 Table 6 shows the results of the ACC (mean$\pm$std) of each contrastive algorithm on each UCI dataset and the average ACC ($\overline{ACC}$) of each algorithm is listed in the last column. The MC-AE+KM algorithm based on the proposed MC-AE shows the best performance on the balance, biodegradation, dermatology, Kdd, OLD, parkinsons and secom datasets. The ACC of them are 0.6224, 0.6673, 0.4754, 0.9877, 0.9369, 0.8051 and 0.9336, respectively. The MC-AE+SC algorithm based on the proposed MC-AE shows the best performance on the car and HabermanSurvial datasets. The ACC of them are 0.6985 and 0.7255. The average ACC of MC-AE+KM and MC-AE+SC algorithms are 0.7049 and 0.7108, respectively. They show the most excellent performance in the corresponding comparative grouping.  \\
\indent An intuitive comparison of the overall performance (average ACC) is shown in Fig. 7 (left one). From Table 6 and Fig. 7, we can draw the conclusion that the proposed MC-AE shows the better performance than other deep models (DAE, Deep-FS, fGraphDBN, CDL and VGAE) in terms of the ACC metric.
\subsubsection{Jaccard Index}
Table 8 shows the results of the Jac of each contrastive algorithm on each dataset and the average Jac ($\overline{Jac}$) of each algorithm is listed in the last column. The MC-AE+KM algorithm based on the proposed MC-AE shows the best performance on the dermatology, Kdd, OLD and secom datasets. The Jac of them are 0.3555, 0.9806, 0.8816 and 0.7634, respectively. The MC-AE+SC algorithm based on the proposed MC-AE shows the best performance on the balance, biodegradation, car and parkinsons datasets. The Jac of them are 0.4285, 0.5525, 0.5408 and 0.6319, respectively. For the ClimateMode and HabermanSurvival datasets, the fGraphDBN+SC and DAE+KM algorithms show the best performance, respectively. Nevertheless the MC-AE+KM and MC-AE+SC algorithms show the best performance in the corresponding comparative grouping. The average Jac of them are 0.5137 and 0.5602, respectively. \\
\indent The intuitive comparison of overall performance (average Jac) is shown in Fig. 7 (middle one). Therefore, we can draw the conclusion that the proposed MC-AE shows the best performance among all contrastive deep models in terms of the Jac metric from Table 8 and Fig. 7.
\subsubsection{Fowlkes and Mallows Index}
Table 9 shows the results of the FMI of each contrastive algorithm on each UCI dataset and the average FMI ($\overline{FMI}$) of each algorithm is listed in the last column. The MC-AE+KM algorithm based on the proposed MC-AE shows the best performance on the dermatology, Kdd, OLD and secom datasets. The FMI of them are 0.5249, 0.9853, 0.9390 and 0.8658, respectively. The MC-AE+SC algorithm based on the proposed MC-AE shows the best performance on the balance, biodegradation, car and parkinsons datasets. The FMI of them are 0.6533, 0.7429, 0.7346 and 0.7938, respectively. For the ClimateMode and HabermanSurvival datasets, the fGraphDBN+SC and DAE+KM algorithms show the best performance, respectively. Nevertheless, the MC-AE+KM and MC-AE+SC algorithms show the best performance in the corresponding comparative grouping. The average FMI of them are 0.6498 and 0.7315, respectively. \\
\indent An intuitive comparison of average FMI is shown in Fig. 7 (right one). In terms of the FMI metric, we also can draw the conclusion that the proposed MC-AE shows the better performance than DAE, Deep-FS, fGraphDBN, CDL and VGAE deep models from Fig. 7 and Table 9.
\subsubsection{The Friedman Aligned Ranks Test Statistic}
Table 7 shows the ranks (in the parentheses) and average ranks of all contrastive algorithms. The smaller rank means the better performance of the algorithm on the corresponding dataset. The average ranks of MC-AE+SC and MC-AE+KM algorithms based on our MC-AE are 29.8.7 and 35.5, respectively. However, the average ranks of DAE+KM, Deep-FS+KM, fGraphDBN+KM, CDL+KM and VGAE+KM are 71.7, 84.5, 74.75, 78.1 and 78.4, respectively. And the ranks of DAE+SC, Deep-FS+SC, fGraphDBN+SC, CDL+SC and VGAE+SC are 47.8, 49.7, 60.45, 59.7 and 54.6, respectively. Clearly, MC-AE+SC and MC-AE+KM algorithms show the most excellent performance in the corresponding comparative grouping. By means of the Friedman Aligned test statistic, \emph{T}=7.2492 is the chi-square distribution with 11 degrees of freedom. The \emph{p}-value is $0.00000823$ which is computed by $\chi^2(11)$ distribution for one tailed test and two-tailed probability is 0.00001646. Then, the null hypothesis is rejected at a high level significance. The \emph{p}-values are far less than 0.05, so the experimental results of algorithms are different.\\
\subsubsection{Friedman + Post-hoc Nemenyi Tests}
The results of Friedman test + post-hoc Nemenyi test \cite{2021A} are shown in Fig. 6 among all contrastive methods on UCI datasets. It is obvious that the test values of MC-AE+KM based on our MC-AE model versus DAE+KM, Deep-FS+KM, fGraphDBN+KM, CDL+KM and VGAE+KM are less than 0.05. Hence, there are striking differences between MC-AE+KM and five related contrastive methods (DAE+KM, Deep-FS+KM, fGraphDBN+KM, CDL+KM and VGAE+KM). In Fig. 6, the test values between our MC-AE+SC and five related contrastive methods (DAE+SC Deep-FS+SC, fGraphDBN+SC, CDL+SC and VGAE+SC) are 0.4865, 0.2108, 0.0852, 0.1224, 0.4422, respectively. Although the MC-AE+SC method based on our MC-AE model has better performance than five related contrastive methods, there are no significant difference between MC-AE+SC and them.
\begin{figure*}
\vspace{0.5mm} \centering
    \includegraphics[scale=0.6025]{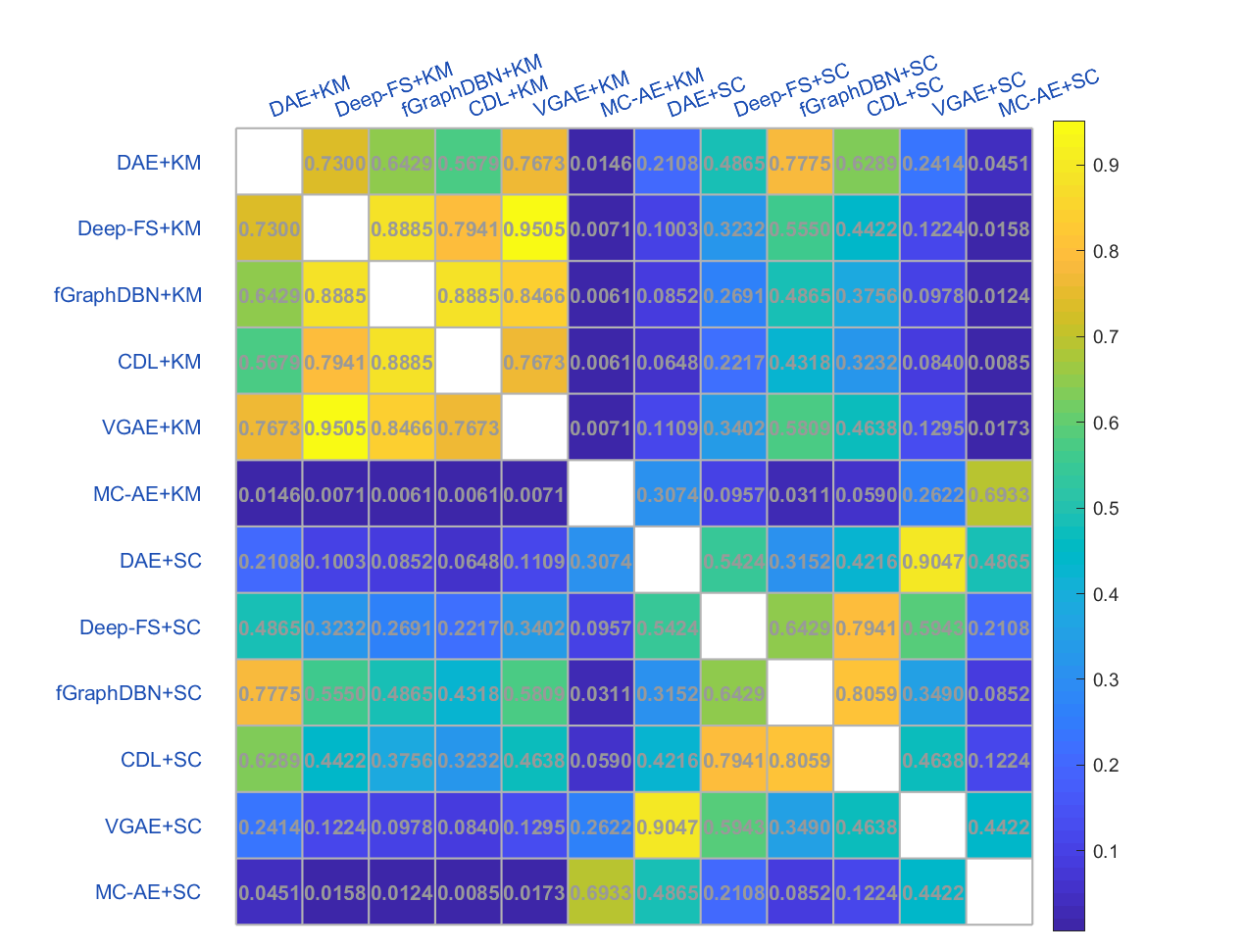}
\\
\caption{The results of Friedman + post-hoc Nemenyi tests among all contrastive algorithms on UCI datasets..
} \label{fig:1}
\end{figure*}
\begin{figure*}
\vspace{0.5mm} \centering
   \includegraphics[scale=0.269025]{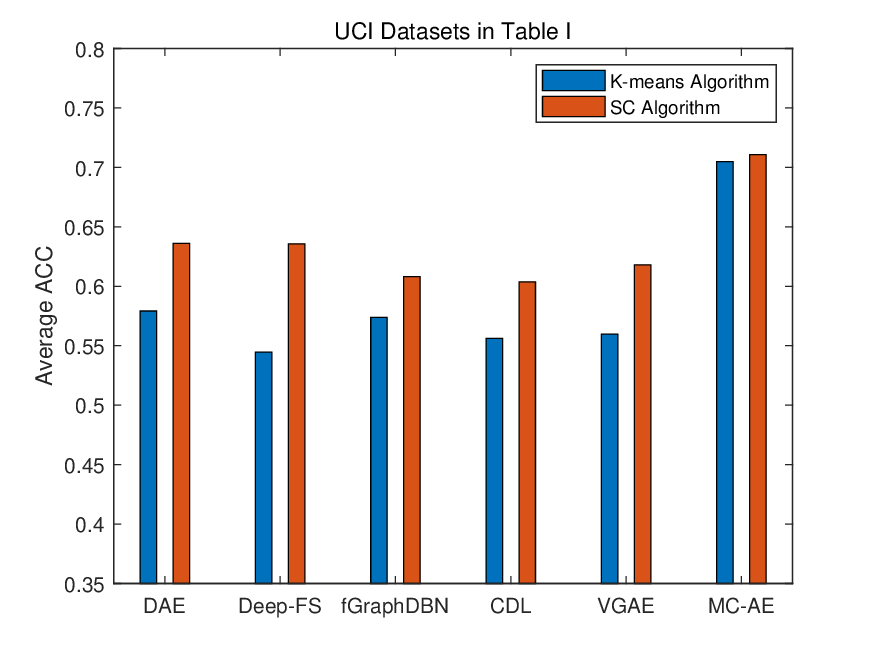}
    \includegraphics[scale=0.269025]{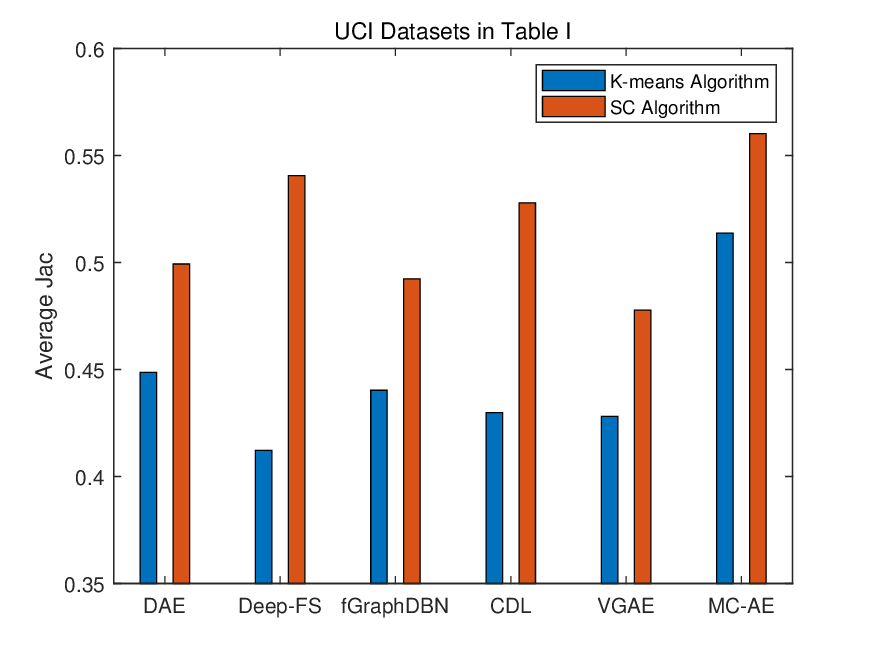}
    \includegraphics[scale=0.269025]{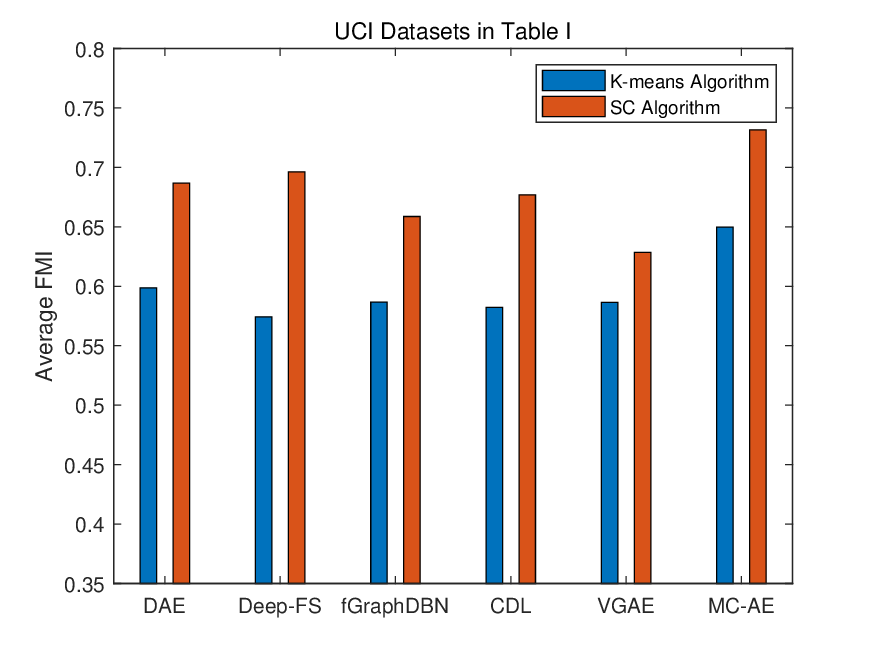}
\\
\caption{The performance comparisons of DAE, Deep-FS, fGraphDBN, CDL, VGAE and our MC-AE using average ACC, Jac and FMI metrics on the UCI datasets.
} \label{fig:1}
\end{figure*}
\subsection{Computational Efforts}
The results of computational efforts (CPU time) of our MC-AE+KM and MC-AE+SC methods on real-valued datasets and UCI datasets are listed in Table 12 and Table 13, respectively. The CPU times of MC-AE+KM algorithm consists of the training time of MC-AE model and clustering time of K-means algorithm. Similarly, the CPU times of MC-AE+SC algorithm consists of the training time of MC-AE model and clustering time of SC algorithm. It is obvious that the training time of MC-AE model occupies the most CPU times of MC-AE+KM and MC-AE+SC algorithms, especially on the high-dimensional datasets. The result was not unexpected because the more local collaborative blocks participate in training process of our MC-AE model on the high-dimensional datasets.
\section{Conclusions}
In this study, we developed a new Multi-local Collaborative AutoEncoder architecture, MC-AE, which is based on the proposed novel mcrRBM and mcrGRBM shallow models. The structure and multi-local collaborative relationships of unlabeled data are integrated into the encoding procedure of our MC-AE that force the multi-local hidden features to converge on the their centers of each local collaborative block. The proposed MC-AE is evaluated on ten real-valued datasets and ten UCI datasets with linear and binary visible layer units, respectively. Through extensive experiments, our MC-AE has consistently outperformed the existing related deep models. Furthermore, the proposed architecture showed more excellent generalization capability for different unsupervised clustering algorithms. In the future work, it is necessary to study multi-local collaborative representation learning for semi-supervised clustering, classification and computer vision.
\section{Acknowledgement}
This work was supported by the National Natural Science Foundation of China (No. 62176221, 71901158) and Sichuan Science and Technology Program (2021YFS0178).



\section{Appendix}
\begin{sidewaystable}[h]
\begin{center}
\caption{The results of the ACC between our MC-AE and the contrastive deep models on ten real-valued datasets. The best performance on each data set is bolded.}
\label{tab:results1} \scalebox{0.63}{
\renewcommand{\arraystretch}{1.5}
\begin{tabular}{cccccccccccc}
\toprule[1.5pt] \textsf{ \bf{\textcolor[rgb]{0.00,0.25,0.50}{Methods}}} & 			{\textbf{banner}} 			&	{\textbf{beret}}			&	{\textbf{bugatti}}			&	{\textbf{building}}			&	{\textbf{vista}} 			&	 {\textbf{vistawallpaper}}			&	{\textbf{voituretuning}} 			&	 {\textbf{water}}			&	 {\textbf{wing}}			&	 {\textbf{worldmap}}			&	 {\textbf{$\textcolor[rgb]{0.00,0.00,0.63}{\overline{ACC}}$}}	\\																		
\toprule[1.5pt]\textsf{	\textcolor[rgb]{0.00,0.25,0.50}{DAE+KM}	}&{	0.5105 	$\pm$	0.0004 	}&{	0.3893 	$\pm$	0.0018 	}&{	0.4161 	$\pm$	0.0010 	}&{	0.4501 	$\pm$	0.0013 	}&{	0.3842 	$\pm$	0.0006 	}&{	0.3780 	$\pm$	0.0008 	}&{	0.3788 	$\pm$	0.0013 	}&{	0.3482 	$\pm$	0.0009 	}&{	0.3680 	$\pm$	0.0010 	}&{	0.4834 	$\pm$	0.0020 	}&{	\textcolor[rgb]{0.00,0.00,0.63}{0.4107} 	}\\																		
\textsf{	\textcolor[rgb]{0.00,0.25,0.50}{Deep-FS+KM}	}&{	0.4364 	$\pm$	0.0003 	}&{	0.4254 	$\pm$	0.0000 	}&{	0.4305 	$\pm$	0.0005 	}&{	0.5159 	$\pm$	0.0001 	}&{	0.4251 	$\pm$	0.0003 	}&{	0.4251 	$\pm$	0.0003 	}&{	0.4331 	$\pm$	0.0002 	}&{	0.4140 	$\pm$	0.0000 	}&{	0.4151 	$\pm$	0.0003 	}&{	0.4731 	$\pm$	0.0002 	}&{	\textcolor[rgb]{0.00,0.00,0.63}{0.4394} 	}\\																		
\textsf{	\textcolor[rgb]{0.00,0.25,0.50}{fGraphDBN+KM}	}&{	0.4802 	$\pm$	0.0148 	}&{	0.4361 	$\pm$	0.0010 	}&{	0.3515 	$\pm$	0.0853 	}&{	0.5642 	$\pm$	0.0011 	}&{	0.4030 	$\pm$	0.0311 	}&{	0.4155 	$\pm$	0.0103 	}&{	0.4892 	$\pm$	0.0014 	}&{	0.4121 	$\pm$	0.0082 	}&{	0.4334 	$\pm$	0.0027 	}&{	0.5209 	$\pm$	0.0091 	}&{	\textcolor[rgb]{0.00,0.00,0.63}{0.4506} 	}\\																		
\textsf{	\textcolor[rgb]{0.00,0.25,0.50}{CDL+KM}	}&{	0.5267 	$\pm$	0.0000 	}&{	0.4304 	$\pm$	0.0001 	}&{	0.4390 	$\pm$	0.0000 	}&{	0.4918 	$\pm$	0.0014 	}&{	0.4030 	$\pm$	0.0000 	}&{	0.4030 	$\pm$	0.0000 	}&{	0.4020 	$\pm$	0.0001 	}&{	0.3955 	$\pm$	0.0002 	}&{	0.4065 	$\pm$	0.0000 	}&{	0.5077 	$\pm$	0.0001 	}&{	\textcolor[rgb]{0.00,0.00,0.63}{0.4406} 	}\\																		
\textsf{	\textcolor[rgb]{0.00,0.25,0.50}{VGAE+KM	}}&{	0.4905 	$\pm$	0.0237 	}&{	0.4410 	$\pm$	0.0188 	}&{	0.3875 	$\pm$	0.0242 	}&{	0.5463 	$\pm$	0.0343 	}&{	0.4750 	$\pm$	0.0198 	}&{	0.4707 	$\pm$	0.0176 	}&{	0.3985 	$\pm$	0.0141 	}&{	0.3830 	$\pm$	0.0329 	}&{	0.4311 	$\pm$	0.0444 	}&{	0.3969 	$\pm$	0.0227 	}&{	\textcolor[rgb]{0.00,0.00,0.63}{0.4420} 	}\\																		
\hline\textsf{	\textbf{\textcolor[rgb]{0.00,0.25,0.50}{MC-AE+KM}}	}&{	\textbf{0.9372 	$\pm$	0.0006} 	}&{	\textbf{0.6895 	$\pm$	0.0002} 	}&{	0.4989 	$\pm$	0.0003 	}&{	\textbf{0.7146 	$\pm$	0.0000} 	}&{	\textbf{0.6308 	$\pm$	0.0005} 	}&{	0.4906 	$\pm$	0.0002 	}&{	\textbf{0.6394 	$\pm$	0.0009} 	}&{	0.5108 	$\pm$	0.0004 	}&{	\textbf{0.6192 	$\pm$	0.0053} 	}&{	0.6043 	$\pm$	0.0030 	}&{\textbf{	\textcolor[rgb]{0.00,0.00,0.63}{0.6335} }	}\\																		
\hline\textsf{	\textcolor[rgb]{0.00,0.25,0.50}{DAE+SC}	}&{	0.3570 	$\pm$	0.0000 	}&{	0.3642 	$\pm$	0.0001 	}&{	0.3639 	$\pm$	0.0002 	}&{	0.3469 	$\pm$	0.0001 	}&{	0.3567 	$\pm$	0.0001 	}&{	0.3717 	$\pm$	0.0002 	}&{	0.3481 	$\pm$	0.0001 	}&{	0.3449 	$\pm$	0.0001 	}&{	0.3540 	$\pm$	0.0001 	}&{	0.3455 	$\pm$	0.0001 	}&{\textcolor[rgb]{0.00,0.00,0.63}{	0.3553} 	}\\																		
\textsf{	\textcolor[rgb]{0.00,0.25,0.50}{Deep-FS+SC	}}&{	0.5093 	$\pm$	0.0001 	}&{	0.4001 	$\pm$	0.0000 	}&{	0.4947 	$\pm$	0.0001 	}&{	0.4812 	$\pm$	0.0030 	}&{	0.4927 	$\pm$	0.0000 	}&{	0.4927 	$\pm$	0.0002 	}&{	0.4391 	$\pm$	0.0000 	}&{	0.4638 	$\pm$	0.0001 	}&{	0.5456 	$\pm$	0.0001 	}&{	0.5672 	$\pm$	0.0000 	}&{	\textcolor[rgb]{0.00,0.00,0.63}{0.4886 }	}\\																		
\textsf{	\textcolor[rgb]{0.00,0.25,0.50}{fGraphDBN+SC}	}&{	0.4035 	$\pm$	0.0013 	}&{	0.4132 	$\pm$	0.0412 	}&{	0.3662 	$\pm$	0.0021 	}&{	0.5269 	$\pm$	0.0010 	}&{	0.4618 	$\pm$	0.0063 	}&{	0.4718 	$\pm$	0.0014 	}&{	0.5154 	$\pm$	0.0021 	}&{	0.4056 	$\pm$	0.0019 	}&{	0.4871 	$\pm$	0.0024 	}&{	0.4824 	$\pm$	0.0103 	}&{\textcolor[rgb]{0.00,0.00,0.63}{	0.4534 }	}\\																		
\textsf{	\textcolor[rgb]{0.00,0.25,0.50}{CDL+SC}	}&{	0.6031 	$\pm$	0.0000 	}&{	0.4063 	$\pm$	0.0000 	}&{	0.4769 	$\pm$	0.0051 	}&{	0.4877 	$\pm$	0.0002 	}&{	0.5424 	$\pm$	0.0053 	}&{	0.5924 	$\pm$	0.0053 	}&{	0.5631 	$\pm$	0.0074 	}&{	0.5201 	$\pm$	0.0000 	}&{	0.5668 	$\pm$	0.0002 	}&{	0.5455 	$\pm$	0.0001 	}&{	\textcolor[rgb]{0.00,0.00,0.63}{0.5304} 	}\\																		
\textsf{	\textcolor[rgb]{0.00,0.25,0.50}{VGAE+SC}	}&{	0.7672 	$\pm$	0.0549 	}&{	0.5696 	$\pm$	0.0104 	}&{	0.6604 	$\pm$	0.0037 	}&{	0.5950 	$\pm$	0.0224 	}&{	0.5111 	$\pm$	0.0579 	}&{	0.4631 	$\pm$	0.0282 	}&{	0.5354 	$\pm$	0.0575 	}&{	0.4086 	$\pm$	0.0203 	}&{	0.5550 	$\pm$	0.0447 	}&{	0.5388 	$\pm$	0.0897 	}&{	\textcolor[rgb]{0.00,0.00,0.63}{0.5604 }	}\\																		
\hline\textsf{	\textbf{\textcolor[rgb]{0.00,0.25,0.50}{MC-AE+SC}}	}&{	0.9349 	$\pm$	0.0089 	}&{	0.4224 	$\pm$	0.0112 	}&{	\textbf{0.7007 	$\pm$	0.0089} 	}&{	0.7058 	$\pm$	0.0026 	}&{	0.5720 	$\pm$	0.0068 	}&{	\textbf{0.6320 	$\pm$	0.0054 }	}&{	0.6359 	$\pm$	0.0025 	}&{	\textbf{0.5705 	$\pm$	0.0044} 	}&{	0.6121 	$\pm$	0.0046 	}&{\textbf{	0.7134 	$\pm$	0.0081 	}}&{\textbf{	\textcolor[rgb]{0.00,0.00,0.63}{0.6500}} 	}\\																									
\hline
\end{tabular}}
\end{center}
\end{sidewaystable}
\begin{sidewaystable}[h]
\begin{center}
\caption{The results of the Rank between our MC-AE and the contrastive deep models on ten real-valued datasets. The smaller the better.}
\label{tab:results1} \scalebox{0.63}{
\begin{tabular}{ccccccccccccc}
\toprule[1.5pt] \textsf{ \bf{\textcolor[rgb]{0.00,0.25,0.50}{Methods}}} & 			{\textbf{banner}} 			&	{\textbf{beret}}			&	{\textbf{bugatti}}			&	{\textbf{building}}			&	{\textbf{vista}} 			&	 {\textbf{vistawallpaper}}			&	{\textbf{voituretuning}} 			&	 {\textbf{water}}			&	 {\textbf{wing}}			&	 {\textbf{worldmap}}			&	 {\textbf{Total}}&{\textcolor[rgb]{0.00,0.00,0.63}{\textbf{Avg}}}	\\	
\toprule[1.5pt] \textsf{	\textcolor[rgb]{0.00,0.25,0.50}{		DAE+KM	}}&{	-0.0692 	(	92 	)}&{	-0.0597 	(	87 	)}&{	-0.0494 	(	82 	)}&{	-0.0854 	(	100 	)}&{	-0.0873 	(	102 	)}&{	-0.0892 	(	104 	)}&{	-0.1027 	(	109 	)}&{	-0.0832 	(	98 	)}&{	-0.1148 	(	112 	)}&{	-0.0315 	(	66 	)}&{	952	}&{	\textcolor[rgb]{0.00,0.00,0.63}{95.2}	}\\
\textsf{	\textcolor[rgb]{0.00,0.25,0.50}{		Deep-FS+KM	}}&{	-0.1433 	(	116 	)}&{	-0.0236 	(	62 	)}&{	-0.0350 	(	68 	)}&{	-0.0196 	(	60 	)}&{	-0.0464 	(	76 	)}&{	-0.0421 	(	72 	)}&{	-0.0484 	(	78 	)}&{	-0.0174 	(	57 	)}&{	-0.0677 	(	89 	)}&{	-0.0418 	(	71 	)}&{	749	}&{	\textcolor[rgb]{0.00,0.00,0.63}{74.9}	}\\
\textsf{	\textcolor[rgb]{0.00,0.25,0.50}{		fGraphDBN+KM	}}&{	-0.0995 	(	107 	)}&{	-0.0129 	(	56 	)}&{	-0.1140 	(	110 	)}&{	0.0287 	(	37 	)}&{	-0.0685 	(	90.5 	)}&{	-0.0517 	(	83 	)}&{	0.0077 	(	45 	)}&{	-0.0193 	(	59 	)}&{	-0.0494 	(	81 	)}&{	0.0060 	(	46 	)}&{	714.5	}&{	\textcolor[rgb]{0.00,0.00,0.63}{71.45}	}\\
\textsf{	\textcolor[rgb]{0.00,0.25,0.50}{		CDL+KM	}}&{	-0.0530 	(	85 	)}&{	-0.0186 	(	58 	)}&{	-0.0265 	(	64 	)}&{	-0.0437 	(	75 	)}&{	-0.0685 	(	90.5 	)}&{	-0.0642 	(	88 	)}&{	-0.0795 	(	96 	)}&{	-0.0359 	(	70 	)}&{	-0.0763 	(	94 	)}&{	-0.0072 	(	52 	)}&{	772.5	}&{	\textcolor[rgb]{0.00,0.00,0.63}{77.25}	}\\
\textsf{	\textcolor[rgb]{0.00,0.25,0.50}{		VGAE+KM	}}&{	-0.0892 	(	103 	)}&{	-0.0080 	(	53 	)}&{	-0.0780 	(	95 	)}&{	0.0108 	(	44 	)}&{	0.0035 	(	49 	)}&{	0.0035 	(	50 	)}&{	-0.0830 	(	97 	)}&{	-0.0484 	(	79 	)}&{	-0.0517 	(	84 	)}&{	-0.1180 	(	113 	)}&{	767	}&{	\textcolor[rgb]{0.00,0.00,0.63}{76.7}	}\\
\hline \textsf{	\textcolor[rgb]{0.00,0.25,0.50}{		\textbf{MC-AE+KM}	}}&{	0.3575 	(	1 	)}&{	0.2405 	(	3 	)}&{	0.0334 	(	33 	)}&{	0.1791 	(	8 	)}&{	0.1593 	(	11 	)}&{	0.0234 	(	41 	)}&{	0.1579 	(	12 	)}&{	0.0794 	(	24 	)}&{	0.1364 	(	15 	)}&{	0.0894 	(	20 	)}&{	\textbf{168}	}&{	\textbf{\textcolor[rgb]{0.00,0.00,0.63}{16.8}}	}\\
\hline \textsf{	\textcolor[rgb]{0.00,0.25,0.50}{		DAE+SC	}}&{	-0.2227 	(	120 	)}&{	-0.0848 	(	99 	)}&{	-0.1016 	(	108 	)}&{	-0.1886 	(	119 	)}&{	-0.1148 	(	111 	)}&{	-0.0955 	(	105 	)}&{	-0.1334 	(	115 	)}&{	-0.0865 	(	101 	)}&{	-0.1288 	(	114 	)}&{	-0.1694 	(	117 	)}&{	1109	}&{	\textcolor[rgb]{0.00,0.00,0.63}{110.9}	}\\
\textsf{	\textcolor[rgb]{0.00,0.25,0.50}{		Deep-FS+SC	}}&{	-0.0704 	(	93 	)}&{	-0.0489 	(	80 	)}&{	0.0292 	(	36 	)}&{	-0.0543 	(	86 	)}&{	0.0212 	(	42 	)}&{	0.0255 	(	38 	)}&{	-0.0424 	(	73 	)}&{	0.0324 	(	34 	)}&{	0.0628 	(	27 	)}&{	0.0523 	(	30 	)}&{	539	}&{	\textcolor[rgb]{0.00,0.00,0.63}{53.9}	}\\
\textsf{	\textcolor[rgb]{0.00,0.25,0.50}{		fGraphDBN+SC	}}&{	-0.1762 	(	118 	)}&{	-0.0358 	(	69 	)}&{	-0.0993 	(	106 	)}&{	-0.0086 	(	54 	)}&{	-0.0097 	(	55 	)}&{	0.0046 	(	47 	)}&{	0.0339 	(	32 	)}&{	-0.0258 	(	63 	)}&{	0.0043 	(	48 	)}&{	-0.0325 	(	67 	)}&{	659	}&{\textcolor[rgb]{0.00,0.00,0.63}{	65.9}	}\\
\textsf{	\textcolor[rgb]{0.00,0.25,0.50}{		CDL+SC	}}&{	0.0234 	(	40 	)}&{	-0.0427 	(	74 	)}&{	0.0114 	(	43 	)}&{	-0.0478 	(	77 	)}&{	0.0709 	(	26 	)}&{	0.1252 	(	17 	)}&{	0.0816 	(	23 	)}&{	0.0887 	(	21 	)}&{	0.0840 	(	22 	)}&{	0.0306 	(	35 	)}&{	378	}&{\textcolor[rgb]{0.00,0.00,0.63}{	37.8}	}\\
\textsf{	\textcolor[rgb]{0.00,0.25,0.50}{		VGAE+SC	}}&{	0.1875 	(	7 	)}&{	0.1206 	(	18 	)}&{	0.1949 	(	6 	)}&{	0.0595 	(	28 	)}&{	0.0396 	(	31 	)}&{	-0.0041 	(	51 	)}&{	0.0539 	(	29 	)}&{	-0.0228 	(	61 	)}&{	0.0722 	(	25 	)}&{	0.0239 	(	39 	)}&{	295	}&{	\textcolor[rgb]{0.00,0.00,0.63}{29.5}	}\\
\hline \textsf{	\textcolor[rgb]{0.00,0.25,0.50}{		\textbf{MC-AE+SC}	}}&{	0.3552 	(	2 	)}&{	-0.0266 	(	65 	)}&{	0.2352 	(	4 	)}&{	0.1703 	(	9 	)}&{	0.1005 	(	19 	)}&{	0.1648 	(	10 	)}&{	0.1544 	(	13 	)}&{	0.1391 	(	14 	)}&{	0.1293 	(	16 	)}&{	0.1985 	(	5 	)}&{	\textbf{157}	}&{	\textbf{\textcolor[rgb]{0.00,0.00,0.63}{15.7}}	}\\
\hline \textsf{	\textcolor[rgb]{0.00,0.25,0.50}{		Total	}}&{			884	}&{			724	}&{			755	}&{			697	}&{			703	}&{			706	}&{			722	}&{			681	}&{			727	}&{			661	}&{	7260	}&{}\\				
\hline
\end{tabular}}
\end{center}
\end{sidewaystable}

\begin{sidewaystable}[h]
\begin{center}
\caption{The results of the Jac between our MC-AE and the contrastive deep models on ten real-valued datasets. The best performance on each data set is bolded.}
\label{tab:results1} \scalebox{0.63}{
\begin{tabular}{cccccccccccc}
\toprule[1.5pt]
 \textsf{ \bf{\textcolor[rgb]{0.00,0.25,0.50}{Methods}}} & 			{\textbf{banner}} 			&	{\textbf{beret}}			&	{\textbf{bugatti}}			&	{\textbf{building}}			&	{\textbf{vista}} 			&	 {\textbf{vistawallpaper}}			&	{\textbf{voituretuning}} 			&	 {\textbf{water}}			&	 {\textbf{wing}}			&	 {\textbf{worldmap}}			&	 {\textbf{$\textcolor[rgb]{0.00,0.00,0.63}{\overline{Jac}}$}}\\	
\toprule[1.5pt]\textsf{ \textcolor[rgb]{0.00,0.25,0.50}{DAE+KM}}&{	0.3720 			}&{	0.2825 			}&{	0.2774 			}&{	0.3070 			}&{	0.2582 			}&{	0.2796 			}&{	0.2515 			}&{	0.2321 			}&{	0.2491 			}&{	0.3145 			}&{	\textcolor[rgb]{0.00,0.00,0.63}{0.2824 }	}\\
\textsf{ \textcolor[rgb]{0.00,0.25,0.50}{Deep-FS+KM}}&{	0.3363 			}&{	0.2755 			}&{	0.2792 			}&{	0.3480 			}&{	0.2780 			}&{	0.2780 			}&{	0.2765 			}&{	0.2495 			}&{	0.2727 			}&{	0.3181 			}&{	\textcolor[rgb]{0.00,0.00,0.63}{0.2912} 	}\\
\textsf{ \textcolor[rgb]{0.00,0.25,0.50}{fGraphDBN+KM}}&{	0.3561 			}&{	0.2900 			}&{	0.2614 			}&{	0.3880 			}&{	0.2708 			}&{	0.2677 			}&{	0.3117 			}&{	0.2706 			}&{	0.2744 			}&{	0.3490 			}&{	\textcolor[rgb]{0.00,0.00,0.63}{0.3040 }	}\\
\textsf{ \textcolor[rgb]{0.00,0.25,0.50}{CDL+KM}}&{	0.3955 			}&{	0.3082 			}&{	0.3090 			}&{	0.3406 			}&{	0.2751 			}&{	0.2751 			}&{	0.2584 			}&{	0.2473 			}&{	0.2663 			}&{	0.3555 			}&{	\textcolor[rgb]{0.00,0.00,0.63}{0.3031} 	}\\
\textsf{ \textcolor[rgb]{0.00,0.25,0.50}{VGAE+KM}}&{	0.3613 			}&{	0.2886 			}&{	0.2731 			}&{	0.3397 			}&{	0.2832 			}&{	0.2814 			}&{	0.2589 			}&{	0.2393 			}&{	0.2732 			}&{	0.2809 			}&{	\textcolor[rgb]{0.00,0.00,0.63}{0.2880} 	}\\
\hline\textsf{ \textcolor[rgb]{0.00,0.25,0.50}{\textbf{MC-AE+KM}}}&{	\textbf{0.8820} 			}&{	\textbf{0.5348 	}		}&{	0.3194 			}&{	\textbf{0.5574 }			}&{	\textbf{0.4738 	}		}&{	0.2714 			}&{	\textbf{0.4760} 			}&{	0.3561 			}&{\textbf{	0.4714 }			}&{	0.4290 			}&{	\textbf{\textcolor[rgb]{0.00,0.25,0.50}{0.4771} }	}\\
\hline\textsf{ \textcolor[rgb]{0.00,0.25,0.50}{DAE+SC}}&{	0.3192 			}&{	0.2633 			}&{	0.2606 			}&{	0.2629 			}&{	0.2435 			}&{	0.2432 			}&{	0.2435 			}&{	0.2321 			}&{	0.2430 			}&{	0.2647 			}&{	\textcolor[rgb]{0.00,0.00,0.63}{0.2576 }	}\\
\textsf{ \textcolor[rgb]{0.00,0.25,0.50}{Deep-FS+SC}}&{	0.4672 			}&{	0.3496 			}&{	0.3530 			}&{	0.2717 			}&{	0.3314 			}&{	0.3314 			}&{	0.3236 			}&{	0.3085 			}&{	0.3433 			}&{	0.3885 			}&{	\textcolor[rgb]{0.00,0.00,0.63}{0.3468} 	}\\
\textsf{ \textcolor[rgb]{0.00,0.25,0.50}{fGraphDBN+SC}}&{	0.3441 			}&{	0.2750 			}&{	0.2624 			}&{	0.3502 			}&{	0.3226 			}&{	0.3221 			}&{	0.3432 			}&{	0.2603 			}&{	0.3305 			}&{	0.3272 			}&{	\textcolor[rgb]{0.00,0.00,0.63}{0.3138} 	}\\
\textsf{ \textcolor[rgb]{0.00,0.25,0.50}{CDL+SC}}&{	0.4906 			}&{	0.3679 			}&{	0.2694 			}&{	0.3657 			}&{	0.3318 			}&{	0.3318 			}&{	0.3271 			}&{	0.4351 			}&{	0.4694 			}&{	0.3887 			}&{	\textcolor[rgb]{0.00,0.00,0.63}{0.3777} 	}\\
\textsf{ \textcolor[rgb]{0.00,0.25,0.50}{VGAE+SC}}&{	0.6274 			}&{	0.4076 			}&{	0.5046 			}&{	0.4096 			}&{	0.3759 			}&{	0.3464 			}&{	0.3706 			}&{	0.2589 			}&{	0.3893 			}&{	0.3924 			}&{	\textcolor[rgb]{0.00,0.00,0.63}{0.4083 }	}\\
\hline\textsf{ \bf{\textcolor[rgb]{0.00,0.25,0.50}{\textbf{MC-AE+SC}}}}&{	0.8779 			}&{	0.5328 			}&{	\textbf{0.5420 }			}&{	0.5470 			}&{	0.4120 			}&{	\textbf{0.4736 	}		}&{	0.4709 			}&{	\textbf{0.4351} 			}&{	0.4598 			}&{	\textbf{0.5602 }			}&{	\textbf{\textcolor[rgb]{0.00,0.25,0.50}{0.5311}} 	}\\
\hline
\end{tabular}}
\end{center}
\end{sidewaystable}

\begin{sidewaystable}[h]
\begin{center}
\caption{The results of the FMI between our MC-AE and the contrastive deep models on ten real-valued datasets. The best performance on each data set is bolded.}
\label{tab:results1} \scalebox{0.63}{
\renewcommand{\arraystretch}{1.5}
\begin{tabular}{cccccccccccc}
\toprule[1.5pt] \textsf{ \bf{\textcolor[rgb]{0.00,0.25,0.50}{Methods}}} & 			{\textbf{banner}} 			&	{\textbf{beret}}			&	{\textbf{bugatti}}			&	{\textbf{building}}			&	{\textbf{vista}} 			&	 {\textbf{vistawallpaper}}			&	{\textbf{voituretuning}} 			&	 {\textbf{water}}			&	 {\textbf{wing}}			&	 {\textbf{worldmap}}			&	 {\textbf{$\textcolor[rgb]{0.00,0.25,0.50}{\overline{FMI}}$}}	\\	
\toprule[1.5pt]\textsf{	\textcolor[rgb]{0.00,0.25,0.50}{DAE+KM}	}&{	0.5876 			}&{	0.4488 			}&{	0.4430 			}&{	0.4765 			}&{	0.4143 			}&{	0.4380 			}&{	0.4065 			}&{	0.3802 			}&{	0.4038 			}&{	0.4849 			}&{\textcolor[rgb]{0.00,0.00,0.63}{	0.4484 }	}\\
\textsf{	\textcolor[rgb]{0.00,0.25,0.50}{Deep-FS+KM	}}&{	0.5577 			}&{	0.4410 			}&{	0.4467 			}&{	0.5223 			}&{	0.4386 			}&{	0.4386 			}&{	0.4361 			}&{	0.4016 			}&{	0.4330 			}&{	0.4915 			}&{	\textcolor[rgb]{0.00,0.00,0.63}{0.4607 }	}\\
\textsf{	\textcolor[rgb]{0.00,0.25,0.50}{fGraphDBN+KM	}}&{	0.5727 			}&{	0.4558 			}&{	0.4261 			}&{	0.5591 			}&{	0.4280 			}&{	0.4248 			}&{	0.4755 			}&{	0.4259 			}&{	0.4327 			}&{	0.5188 			}&{	\textcolor[rgb]{0.00,0.00,0.63}{0.4719} 	}\\
\textsf{	\textcolor[rgb]{0.00,0.25,0.50}{CDL+KM}	}&{	0.6070 			}&{	0.4771 			}&{	0.4778 			}&{	0.5123 			}&{	0.4338 			}&{	0.4338 			}&{	0.4160 			}&{	0.3986 			}&{	0.4256 			}&{	0.5266 			}&{	\textcolor[rgb]{0.00,0.00,0.63}{0.4709} 	}\\
\textsf{	\textcolor[rgb]{0.00,0.25,0.50}{VGAE+KM}	}&{	0.5764 			}&{	0.4550 			}&{	0.4382 			}&{	0.5144 			}&{	0.4442 			}&{	0.4422 			}&{	0.4156 			}&{	0.3891 			}&{	0.4316 			}&{	0.4493 			}&{	\textcolor[rgb]{0.00,0.00,0.63}{0.4556} 	}\\
\hline\textsf{	\textbf{\textcolor[rgb]{0.00,0.25,0.50}{MC-AE+KM}}	}&{	\textbf{0.9392 }			}&{	\textbf{0.7313 	}		}&{	0.4902 			}&{	\textbf{0.7466} 			}&{	\textbf{0.6870 	}		}&{	0.4321 			}&{	\textbf{0.6899 	}		}&{	0.5344 			}&{	\textbf{0.6866 }			}&{	0.6043 			}&{	\textbf{\textcolor[rgb]{0.00,0.00,0.63}{0.6542} }	}\\
\hline\textsf{	\textcolor[rgb]{0.00,0.25,0.50}{DAE+SC}	}&{	0.5424 			}&{	0.4285 			}&{	0.4259 			}&{	0.4303 			}&{	0.3979 			}&{	0.3975 			}&{	0.3979 			}&{	0.3803 			}&{	0.3969 			}&{	0.4333 			}&{	\textcolor[rgb]{0.00,0.00,0.63}{0.4231 }	}\\
\textsf{	\textcolor[rgb]{0.00,0.25,0.50}{Deep-FS+SC}	}&{	0.6629 			}&{	0.5184 			}&{	0.5223 			}&{	0.4416 			}&{	0.4980 			}&{	0.4980 			}&{	0.4890 			}&{	0.4276 			}&{	0.5113 			}&{	0.5603 			}&{	\textcolor[rgb]{0.00,0.00,0.63}{0.5129 }	}\\
\textsf{	\textcolor[rgb]{0.00,0.25,0.50}{fGraphDBN+SC}	}&{	0.5633 			}&{	0.4401 			}&{	0.4272 			}&{	0.5202 			}&{	0.4880 			}&{	0.4874 			}&{	0.5116 			}&{	0.4133 			}&{	0.4972 			}&{	0.4967 			}&{	\textcolor[rgb]{0.00,0.00,0.63}{0.4845 }	}\\
\textsf{	\textcolor[rgb]{0.00,0.25,0.50}{CDL+SC}	}&{	0.6805 			}&{	0.5380 			}&{	0.4345 			}&{	0.5360 			}&{	0.4984 			}&{	0.4984 			}&{	0.4931 			}&{	0.4585 			}&{	0.6840 			}&{	0.5600 			}&{\textcolor[rgb]{0.00,0.00,0.63}{	0.5381 }	}\\
\textsf{	\textcolor[rgb]{0.00,0.25,0.50}{VGAE+SC}	}&{	0.7756 			}&{	0.5830 			}&{	0.6947 			}&{	0.5831 			}&{	0.5595 			}&{	0.5207 			}&{	0.5487 			}&{	0.4119 			}&{	0.5714 			}&{	0.5638 			}&{	\textcolor[rgb]{0.00,0.00,0.63}{0.5812} 	}\\
\hline\textsf{\textbf{\textcolor[rgb]{0.00,0.25,0.50}{	MC-AE+SC}	}}&{	0.9367 			}&{	0.7288 			}&{	\textbf{0.7279 	}		}&{	0.7353 			}&{	0.6028 			}&{	\textbf{0.6865 	}		}&{	0.6808 			}&{	\textbf{0.6585 }			}&{	0.6692 			}&{\textbf{	0.7460 	}		}&{\textbf{	\textcolor[rgb]{0.00,0.00,0.63}{0.7172} }	}\\
\hline
\end{tabular}}
\end{center}
\end{sidewaystable}

\begin{sidewaystable}[h]
\begin{center}
\caption{The results of the ACC between our MC-AE and the contrastive deep models on ten UCI datasets. The best performance on each data set is bolded.}
\label{tab:results1} \scalebox{0.63}{
\renewcommand{\arraystretch}{1.5}
\begin{tabular}{cccccccccccc}
\toprule[1.5pt] \textsf{ \bf{\textcolor[rgb]{0.00,0.25,0.50}{Methods}}} & 			{\textbf{balance}}			&	{\textbf{biodegradation}}			&	{\textbf{car}}			&	{\textbf{ClimateModel}}			&	{\textbf{dermatology}}			&	{\textbf{HabermanSurvival}}			&	{\textbf{Kdd}}			&	{\textbf{OLD}}			&	{\textbf{parkinsons}}			&	{\textbf{secom}}			&	{\textbf{$\textcolor[rgb]{0.00,0.00,0.63}{\overline{ACC}}$}}	\\								
\toprule[1.5pt]\textsf{	\textcolor[rgb]{0.00,0.25,0.50}{DAE+KM	}}&{	0.3904 	$\pm$	0.0000 	}&{	0.6199 	$\pm$	0.0002 	}&{	0.3414 	$\pm$	0.0000 	}&{	0.5130 	$\pm$	0.0001 	}&{	0.4372 	$\pm$	0.0010 	}&{	0.6242 	$\pm$	0.0009 	}&{	0.5523 	$\pm$	0.0000 	}&{	0.9065 	$\pm$	0.0002 	}&{	0.6359 	$\pm$	0.0002 	}&{	0.7709 	$\pm$	0.0006 	}&{	\textcolor[rgb]{0.00,0.00,0.63}{0.5792} 	}\\								
\textsf{	\textcolor[rgb]{0.00,0.25,0.50}{Deep-FS+KM	}}&{	0.4768 	$\pm$	0.0017 	}&{	0.6398 	$\pm$	0.0000 	}&{	0.4199 	$\pm$	0.0000 	}&{	0.5407 	$\pm$	0.0001 	}&{	0.2377 	$\pm$	0.0000 	}&{	0.6157 	$\pm$	0.0000 	}&{	0.3875 	$\pm$	0.0000 	}&{	0.7616 	$\pm$	0.0000 	}&{	0.6051 	$\pm$	0.0000 	}&{	0.7620 	$\pm$	0.0072 	}&{	\textcolor[rgb]{0.00,0.00,0.63}{0.5447 }	}\\								
\textsf{	\textcolor[rgb]{0.00,0.25,0.50}{fGraphDBN+KM}	}&{	0.4176 	$\pm$	0.0339 	}&{	0.5768 	$\pm$	0.0007 	}&{	0.3163 	$\pm$	0.0012 	}&{	0.8444 	$\pm$	0.0943 	}&{	0.3689 	$\pm$	0.0541 	}&{	0.5686 	$\pm$	0.0139 	}&{	0.3695 	$\pm$	0.0110 	}&{	0.8346 	$\pm$	0.0028 	}&{	0.6795 	$\pm$	0.0036 	}&{	0.7625 	$\pm$	0.0221 	}&{	\textcolor[rgb]{0.00,0.00,0.63}{0.5739 }	}\\								
\textsf{\textcolor[rgb]{0.00,0.25,0.50}{	CDL+KM}	}&{	0.4208 	$\pm$	0.0000 	}&{	0.5223 	$\pm$	0.0000 	}&{	0.4132 	$\pm$	0.0000 	}&{	0.5037 	$\pm$	0.0000 	}&{	0.2705 	$\pm$	0.0002 	}&{	0.5196 	$\pm$	0.0000 	}&{	0.6164 	$\pm$	0.0204 	}&{	0.9017 	$\pm$	0.0000 	}&{	0.5641 	$\pm$	0.0040 	}&{	0.8283 	$\pm$	0.0000 	}&{	\textcolor[rgb]{0.00,0.00,0.63}{0.5561} 	}\\								
\textsf{	\textcolor[rgb]{0.00,0.25,0.50}{VGAE+KM}	}&{	0.4886 	$\pm$	0.0993 	}&{	0.6051 	$\pm$	0.0159 	}&{	0.3510 	$\pm$	0.0432 	}&{	0.5420 	$\pm$	0.0304 	}&{	0.4019 	$\pm$	0.0257 	}&{	0.6101 	$\pm$	0.1009 	}&{	0.3650 	$\pm$	0.0030 	}&{	0.9009 	$\pm$	0.0018 	}&{	0.6764 	$\pm$	0.0936 	}&{	0.6570 	$\pm$	0.0216 	}&{	\textcolor[rgb]{0.00,0.00,0.63}{0.5598 }	}\\								
\hline\textsf{	\textbf{\textcolor[rgb]{0.00,0.25,0.50}{MC-AE+KM}}	}&{\textbf{	0.6224 	$\pm$	0.0001 }	}&{\textbf{	0.6673 	$\pm$	0.0004 	}}&{	0.4426 	$\pm$	0.0001 	}&{	0.5500 	$\pm$	0.0004 	}&{\textbf{	0.4754 	$\pm$	0.0005} 	}&{	0.6275 	$\pm$	0.0012 	}&{	\textbf{0.9877 	$\pm$	0.0007 }	}&{	\textbf{0.9369 	$\pm$	0.0002 	} }&{\textbf{	0.8051 	$\pm$	0.0003} 	}&{\textbf{	0.9336 	$\pm$	0.0003} 	}&{	\textbf{\textcolor[rgb]{0.00,0.00,0.63}{0.7049}} 	}\\								
\hline\textsf{	\textcolor[rgb]{0.00,0.25,0.50}{DAE+SC}	}&{	0.4640 	$\pm$	0.0001 	}&{	0.6645 	$\pm$	0.0000 	}&{	0.5434 	$\pm$	0.0000 	}&{	0.5926 	$\pm$	0.0003 	}&{	0.3552 	$\pm$	0.0007 	}&{	\textbf{0.7255 	$\pm$	0.0000} 	}&{	0.6414 	$\pm$	0.0002 	}&{	0.7932 	$\pm$	0.0001 	}&{	0.7641 	$\pm$	0.0000 	}&{	0.8175 	$\pm$	0.0002 	}&{	\textcolor[rgb]{0.00,0.00,0.63}{0.6361 }	}\\								
\textsf{	\textcolor[rgb]{00.00,0.25,0.50}{Deep-FS+SC	}}&{	0.4584 	$\pm$	0.0000 	}&{	0.6190 	$\pm$	0.0000 	}&{	0.6325 	$\pm$	0.0022 	}&{	0.8093 	$\pm$	0.0000 	}&{	0.2814 	$\pm$	0.0006 	}&{	0.7055 	$\pm$	0.0005 	}&{	0.7211 	$\pm$	0.0009 	}&{	0.8481 	$\pm$	0.0260 	}&{	0.7487 	$\pm$	0.0000 	}&{	0.5333 	$\pm$	0.0168 	}&{	\textcolor[rgb]{0.00,0.00,0.63}{0.6357} 	}\\								
\textsf{	\textcolor[rgb]{0.00,0.25,0.50}{fGraphDBN+SC}	}&{	0.4616 	$\pm$	0.0011 	}&{	0.6062 	$\pm$	0.0623 	}&{	0.5966 	$\pm$	0.1457 	}&{	\textbf{0.9120 	$\pm$	0.0013} 	}&{	0.3538 	$\pm$	0.0097 	}&{	0.6977 	$\pm$	0.0023 	}&{	0.3680 	$\pm$	0.0122 	}&{	0.6434 	$\pm$	0.0036 	}&{	0.6795 	$\pm$	0.0036 	}&{	0.7618 	$\pm$	0.0573 	}&{	\textcolor[rgb]{0.00,0.00,0.63}{0.6081 }	}\\								
\textsf{	\textcolor[rgb]{0.00,0.25,0.50}{CDL+SC	}}&{	0.4524 	$\pm$	0.0000 	}&{	0.6445 	$\pm$	0.0000 	}&{	0.6485 	$\pm$	0.0000 	}&{	0.5185 	$\pm$	0.0509 	}&{	0.3087 	$\pm$	0.0000 	}&{	0.6755 	$\pm$	0.0000 	}&{	0.6586 	$\pm$	0.0000 	}&{	0.8833 	$\pm$	0.0000 	}&{	0.7341 	$\pm$	0.0000 	}&{	0.5124 	$\pm$	0.0000 	}&{	\textcolor[rgb]{0.00,0.00,0.63}{0.6037 }	}\\								
\textsf{	\textcolor[rgb]{0.00,0.25,0.50}{VGAE+SC	}}&{	0.5232 	$\pm$	0.0634 	}&{	0.6269 	$\pm$	0.0165 	}&{	0.4064 	$\pm$	0.0659 	}&{	0.6744 	$\pm$	0.1573 	}&{	0.4557 	$\pm$	0.0643 	}&{	0.6229 	$\pm$	0.0772 	}&{	0.4155 	$\pm$	0.0427 	}&{	0.9224 	$\pm$	0.0123 	}&{	0.7118 	$\pm$	0.0620 	}&{	0.8204 	$\pm$	0.0049 	}&{	\textcolor[rgb]{0.00,0.00,0.63}{0.6180 }	}\\								
\hline\textsf{	\textbf{\textcolor[rgb]{0.00,0.25,0.50}{MC-AE+SC}}	}&{	0.4640 	$\pm$	0.0003 	}&{	0.6645 	$\pm$	0.0002 	}&{	\textbf{0.6985 	$\pm$	0.0002 }	}&{	0.9111 	$\pm$	0.0011 	}&{	0.4508 	$\pm$	0.0004 	}&{	\textbf{0.7255 	$\pm$	0.0007} 	}&{	0.8156 	$\pm$	0.0005 	}&{	0.9013 	$\pm$	0.0001 	}&{	0.7641 	$\pm$	0.0000 	}&{	0.7128 	$\pm$	0.0004 	}&{\textbf{	\textcolor[rgb]{0.00,0.00,0.63}{0.7108}} 	}\\								\hline
\end{tabular}}
\end{center}
\end{sidewaystable}

\begin{sidewaystable}[h]
\begin{center}
\caption{The results of the Rank between our MC-AE and the contrastive deep models on ten UCI datasets. The smaller the better.}
\label{tab:results1} \scalebox{0.63}{
\renewcommand{\arraystretch}{1.5}
\begin{tabular}{ccccccccccccc}
 \toprule[1.5pt] \textsf{ \bf{\textcolor[rgb]{0.00,0.25,0.50}{Methods}}} & 			{\textbf{balance}}			&	{\textbf{biodegradation}}			&	{\textbf{car}}			&	{\textbf{ClimateModel}}			&	{\textbf{dermatology}}			&	{\textbf{HabermanSurvival}}			&	{\textbf{Kdd}}			&	{\textbf{OLD}}			&	{\textbf{parkinsons}}			&	{\textbf{secom}}			&	 {\textbf{Total}}&{\textcolor[rgb]{0.00,0.00,0.63}{\textbf{\textcolor[rgb]{0.00,0.00,0.63}{Avg}}}}	\\	
\toprule[1.5pt]\textsf{	\textcolor[rgb]{0.00,0.25,0.50}{		DAE+KM	}}&{	-0.0796 	(	94 	)}&{	-0.0015 	(	59 	)}&{	-0.1428 	(	109 	)}&{	-0.1463 	(	110 	)}&{	0.0708 	(	25 	)}&{	-0.0190 	(	75 	)}&{	-0.0226 	(	78 	)}&{	0.0537 	(	33 	)}&{	-0.0615 	(	88 	)}&{	0.0315 	(	46 	)}&{	717	}&{	\textcolor[rgb]{0.00,0.00,0.63}{71.7}	}\\			
\textsf{	\textcolor[rgb]{0.00,0.25,0.50}{		Deep-FS+KM	}}&{	0.0068 	(	56 	)}&{	0.0184 	(	53 	)}&{	-0.0643 	(	89 	)}&{	-0.1186 	(	103 	)}&{	-0.1287 	(	105 	)}&{	-0.0275 	(	80 	)}&{	-0.1874 	(	114 	)}&{	-0.0912 	(	97 	)}&{	-0.0923 	(	98 	)}&{	0.0226 	(	50 	)}&{	845	}&{	\textcolor[rgb]{0.00,0.00,0.63}{84.5}	}\\			
\textsf{	\textcolor[rgb]{0.00,0.25,0.50}{		fGraphDBN+KM	}}&{	-0.0524 	(	85 	)}&{	-0.0446 	(	83 	)}&{	-0.1679 	(	113 	)}&{	0.1851 	(	7 	)}&{	0.0025 	(	58.0 	)}&{	-0.0746 	(	92 	)}&{	-0.2054 	(	115 	)}&{	-0.0182 	(	74 	)}&{	-0.0179 	(	72.5 	)}&{	0.0231 	(	48 	)}&{	747.5	}&{	\textcolor[rgb]{0.00,0.00,0.63}{74.75}	}\\			
\textsf{	\textcolor[rgb]{0.00,0.25,0.50}{		CDL+KM	}}&{	-0.0492 	(	84 	)}&{	-0.0991 	(	100 	)}&{	-0.0710 	(	91 	)}&{	-0.1556 	(	111 	)}&{	-0.0959 	(	99.0 	)}&{	-0.1236 	(	104 	)}&{	0.0415 	(	42 	)}&{	0.0489 	(	36 	)}&{	-0.1333 	(	107 	)}&{	0.0889 	(	17 	)}&{	791	}&{	\textcolor[rgb]{0.00,0.00,0.63}{79.1}	}\\			
\textsf{	\textcolor[rgb]{0.00,0.25,0.50}{		VGAE+KM	}}&{	0.0186 	(	52 	)}&{	-0.0163 	(	70 	)}&{	-0.1332 	(	106 	)}&{	-0.1173 	(	102 	)}&{	0.0355 	(	44 	)}&{	-0.0331 	(	81 	)}&{	-0.2099 	(	119 	)}&{	0.0481 	(	38 	)}&{	-0.0210 	(	77 	)}&{	-0.0824 	(	95 	)}&{	784	}&{\textcolor[rgb]{0.00,0.00,0.63}{	78.4}	}\\			
\hline\textsf{	\textcolor[rgb]{0.00,0.25,0.50}{		\textbf{MC-AE+KM}	}}&{	0.1524 	(	9 	)}&{	0.0459 	(	39 	)}&{	-0.0416 	(	82 	)}&{	-0.1093 	(	101 	)}&{	0.1090 	(	14 	)}&{	-0.0157 	(	69 	)}&{	0.4128 	(	1 	)}&{	0.0841 	(	19 	)}&{	0.1077 	(	15 	)}&{	0.1942 	(	6 	)}&{	355	}&{	\textcolor[rgb]{0.00,0.00,0.63}{\textbf{35.5}}	}\\			
\hline\textsf{	\textcolor[rgb]{0.00,0.25,0.50}{		DAE+SC	}}&{	-0.0060 	(	62.5 	)}&{	0.0431 	(	40.5 	)}&{	0.0592 	(	31 	)}&{	-0.0667 	(	90 	)}&{	-0.0112 	(	65 	)}&{	0.0823 	(	21.5 	)}&{	0.0665 	(	29 	)}&{	-0.0596 	(	87 	)}&{	0.0667 	(	27.5 	)}&{	0.0781 	(	24 	)}&{	478	}&{	\textcolor[rgb]{0.00,0.00,0.63}{47.8}	}\\			
\textsf{	\textcolor[rgb]{0.00,0.25,0.50}{		Deep-FS+SC	}}&{	-0.0116 	(	66 	)}&{	-0.0024 	(	60 	)}&{	0.1483 	(	11 	)}&{	0.1500 	(	10 	)}&{	-0.0850 	(	96 	)}&{	0.0623 	(	30 	)}&{	0.1462 	(	12 	)}&{	-0.0047 	(	61 	)}&{	0.0513 	(	35 	)}&{	-0.2061 	(	116 	)}&{	497	}&{	\textcolor[rgb]{0.00,0.00,0.63}{49.7}	}\\			
\textsf{	\textcolor[rgb]{0.00,0.25,0.50}{		fGraphDBN+SC	}}&{	-0.0084 	(	64 	)}&{	-0.0152 	(	68 	)}&{	0.1124 	(	13 	)}&{	0.2527 	(	2 	)}&{	-0.0126 	(	67 	)}&{	0.0545 	(	32 	)}&{	-0.2069 	(	117 	)}&{	-0.2094 	(	118 	)}&{	-0.0179 	(	72.5 	)}&{	0.0224 	(	51 	)}&{	604.5	}&{	\textcolor[rgb]{0.00,0.00,0.63}{60.45	}}\\			
\textsf{	\textcolor[rgb]{0.00,0.25,0.50}{		CDL+SC	}}&{	-0.0176 	(	71 	)}&{	0.0231 	(	49 	)}&{	0.1643 	(	8 	)}&{	-0.1408 	(	108 	)}&{	-0.0577 	(	86 	)}&{	0.0323 	(	45 	)}&{	0.0837 	(	20 	)}&{	0.0305 	(	47 	)}&{	0.0367 	(	43 	)}&{	-0.2270 	(	120 	)}&{	597	}&{	\textcolor[rgb]{0.00,0.00,0.63}{59.7}	}\\			
\textsf{	\textcolor[rgb]{0.00,0.25,0.50}{		VGAE+SC	}}&{	0.0532 	(	34 	)}&{	0.0055 	(	57 	)}&{	-0.0778 	(	93 	)}&{	0.0151 	(	54 	)}&{	0.0893 	(	16 	)}&{	-0.0203 	(	76 	)}&{	-0.1594 	(	112 	)}&{	0.0696 	(	26 	)}&{	0.0144 	(	55 	)}&{	0.0810 	(	23 	)}&{	546	}&{	\textcolor[rgb]{0.00,0.00,0.63}{54.6}	}\\			
\hline\textsf{	\textcolor[rgb]{0.00,0.25,0.50}{		\textbf{MC-AE+SC}	}}&{	-0.0060 	(	62.5 	)}&{	0.0431 	(	40.5 	)}&{	0.2143 	(	5 	)}&{	0.2518 	(	3 	)}&{	0.0844 	(	18 	)}&{	0.0823 	(	21.5 	)}&{	0.2407 	(	4 	)}&{	0.0485 	(	37 	)}&{	0.0667 	(	27.5 	)}&{	-0.0266 	(	79 	)}&{	298	}&{	\textbf{\textcolor[rgb]{0.00,0.00,0.63}{29.8}}	}\\			
\hline\textsf{	\textcolor[rgb]{0.00,0.25,0.50}{		Total	}}&{			740	}&{			719	}&{			751	}&{			801	}&{			693	}&{			727	}&{			763	}&{			673	}&{			718	}&{			675	}&{	7260	}&{}\\											
\hline
\end{tabular}}
\end{center}
\end{sidewaystable}

\begin{sidewaystable}[h]
\begin{center}
\caption{The results of the Jac between our MC-AE and the contrastive deep models on ten UCI datasets. The best performance on each data set is bolded.}
\label{tab:results1} \scalebox{0.63}{
\renewcommand{\arraystretch}{1.5}
\begin{tabular}{cccccccccccc}
\toprule[1.5pt] \textsf{ \bf{\textcolor[rgb]{0.00,0.25,0.50}{Methods}}} & 			{\textbf{balance}}			&	{\textbf{biodegradation}}			&	{\textbf{car}}			&	{\textbf{ClimateModel}}			&	{\textbf{dermatology}}			&	{\textbf{HabermanSurvival}}			&	{\textbf{Kdd}}			&	{\textbf{OLD}}			&	{\textbf{parkinsons}}			&	{\textbf{secom}}			&	{\textbf{$\textcolor[rgb]{0.00,0.00,0.63}{\overline{Jac}}$}}	\\	
\toprule[1.5pt]\textsf{	\textcolor[rgb]{0.00,0.25,0.50}{DAE+KM}	}&{	0.2683 			}&{	0.3893 			}&{	0.3050 			}&{	0.4573 			}&{	0.2239 			}&{	\textbf{0.6081 }			}&{	0.4266 			}&{	0.8215 			}&{	0.4103 			}&{	0.5766 			}&{	\textcolor[rgb]{0.00,0.00,0.63}{0.4487 }	}\\								
\textsf{	\textcolor[rgb]{0.00,0.25,0.50}{Deep-FS+KM}	}&{	0.3036 			}&{	0.5126 			}&{	0.2957 			}&{	0.4582 			}&{	0.1099 			}&{	0.5316 			}&{	0.3385 			}&{	0.6213 			}&{	0.4583 			}&{	0.4926 			}&{	\textcolor[rgb]{0.00,0.00,0.63}{0.4122 }	}\\								
\textsf{	\textcolor[rgb]{0.00,0.25,0.50}{fGraphDBN+KM}	}&{	0.2577 			}&{	0.4353 			}&{	0.2194 			}&{	0.7367 			}&{	0.1883 			}&{	0.4049 			}&{	0.2597 			}&{	0.7180 			}&{	0.5232 			}&{	0.6594 			}&{	\textcolor[rgb]{0.00,0.00,0.63}{0.4403} 	}\\								
\textsf{	\textcolor[rgb]{0.00,0.25,0.50}{CDL+KM}	}&{	0.2585 			}&{	0.3916 			}&{	0.2444 			}&{	0.4573 			}&{	0.1192 			}&{	0.3817 			}&{	0.5040 			}&{	0.8215 			}&{	0.4134 			}&{	0.7061 			}&{\textcolor[rgb]{0.00,0.00,0.63}{	0.4298 }	}\\								
\textsf{	\textcolor[rgb]{0.00,0.25,0.50}{VGAE+KM	}}&{	0.3097 			}&{	0.3991 			}&{	0.2329 			}&{	0.4633 			}&{	0.1983 			}&{	0.4382 			}&{	0.3240 			}&{	0.8215 			}&{	0.4882 			}&{	0.6056 			}&{	\textcolor[rgb]{0.00,0.00,0.63}{0.4281} 	}\\								
\hline\textsf{	\textbf{\textcolor[rgb]{0.00,0.25,0.50}{MC-AE+KM}}	}&{	0.2488 			}&{	0.3898 			}&{	0.2143 			}&{	0.4601 			}&{	\textbf{0.3555} 			}&{	0.4324 			}&{	\textbf{0.9806 }			}&{	\textbf{0.8816 	}		}&{	0.4103 			}&{	\textbf{0.7634 	}		}&{	\textbf{\textcolor[rgb]{0.00,0.00,0.63}{0.5137} }	}\\								
\hline\textsf{	\textcolor[rgb]{0.00,0.25,0.50}{DAE+SC}	}&{	0.4285 			}&{	0.5506 			}&{	0.4204 			}&{	0.4719 			}&{	0.2082 			}&{	0.5983 			}&{	0.5492 			}&{	0.4961 			}&{	\textbf{0.6319} 			}&{	0.6383 			}&{\textcolor[rgb]{0.00,0.00,0.63}{	0.4993 }	}\\								
\textsf{	\textcolor[rgb]{0.00,0.25,0.50}{Deep-FS+SC}	}&{	0.4135 			}&{	0.3715 			}&{	0.4657 			}&{	0.8307 			}&{	0.1748 			}&{	0.5748 			}&{	0.5104 			}&{	0.7387 			}&{	0.6108 			}&{	0.7149 			}&{	\textcolor[rgb]{0.00,0.00,0.63}{0.5406} 	}\\								
\textsf{	\textcolor[rgb]{0.00,0.25,0.50}{fGraphDBN+SC}	}&{	0.4267 			}&{	0.4674 			}&{	0.4480 			}&{\textbf{	0.8392 }			}&{	0.2054 			}&{	0.5393 			}&{	0.2852 			}&{	0.5222 			}&{	0.5232 			}&{	0.6660 			}&{	\textcolor[rgb]{0.00,0.00,0.63}{0.4923 }	}\\								
\textsf{	\textcolor[rgb]{0.00,0.25,0.50}{CDL+SC}	}&{	0.4255 			}&{	0.5495 			}&{	0.5376 			}&{	0.4543 			}&{	0.1942 			}&{	0.5953 			}&{	0.5576 			}&{	0.8724 			}&{	0.6289 			}&{	0.4639 			}&{	\textcolor[rgb]{0.00,0.00,0.63}{0.5279 }	}\\								
\textsf{	\textcolor[rgb]{0.00,0.25,0.50}{VGAE+SC}	}&{	0.3203 			}&{	0.4145 			}&{	0.2663 			}&{	0.5799 			}&{	0.2454 			}&{	0.4543 			}&{	0.3354 			}&{	0.8564 			}&{	0.5525 			}&{	0.7524 			}&{	\textcolor[rgb]{0.00,0.00,0.63}{0.4777} 	}\\								
\hline\textsf{	\textbf{\textcolor[rgb]{0.00,0.25,0.50}{MC-AE+SC}}	}&{	\textbf{0.4285 	}		}&{\textbf{	0.5525	}		}&{	\textbf{0.5408} 			}&{	0.8370 			}&{	0.1987 			}&{	0.5983 			}&{	0.5707 			}&{	0.7560 			}&{\textbf{	0.6319} 			}&{	0.4875 			}&{	\textbf{\textcolor[rgb]{0.00,0.00,0.63}{0.5602}} 	}\\								
\hline
\end{tabular}}
\end{center}
\end{sidewaystable}

\begin{sidewaystable}[h]
\begin{center}
\caption{The results of the FMI between our MC-AE and the contrastive deep models on ten UCI datasets. The best performance on each data set is bolded.}
\label{tab:results1} \scalebox{0.63}{
\renewcommand{\arraystretch}{1.5}
\begin{tabular}{cccccccccccc}
\toprule[1.5pt] \textsf{ \bf{\textcolor[rgb]{0.00,0.25,0.50}{Methods}}} & 			{\textbf{balance}}			&	{\textbf{biodegradation}}			&	{\textbf{car}}			&	{\textbf{ClimateModel}}			&	{\textbf{dermatology}}			&	{\textbf{HabermanSurvival}}			&	{\textbf{Kdd}}			&	{\textbf{OLD}}			&	{\textbf{parkinsons}}			&	{\textbf{secom}}			&	{\textbf{$\textcolor[rgb]{0.00,0.00,0.63}{\overline{FMI}}$}}	\\	
\toprule[1.5pt]\textsf{	\textcolor[rgb]{0.00,0.25,0.50}{DAE+KM}	}&{	0.4263 			}&{	0.5551 			}&{	0.3743 			}&{	0.6491 			}&{	0.3659 			}&{	\textbf{0.7778} 			}&{	0.6097 			}&{	0.9023 			}&{	0.5843 			}&{	0.7419 			}&{	\textcolor[rgb]{0.00,0.00,0.63}{0.5987} 	}\\								
\textsf{	\textcolor[rgb]{0.00,0.25,0.50}{Deep-FS+KM}	}&{	0.4570 			}&{	0.6862 			}&{	0.4587 			}&{	0.6400 			}&{	0.1882 			}&{	0.6868 			}&{	0.5718 			}&{	0.7634 			}&{	0.6188 			}&{	0.6711 			}&{	\textcolor[rgb]{0.00,0.00,0.63}{0.5742} 	}\\								
\textsf{	\textcolor[rgb]{0.00,0.25,0.50}{fGraphDBN+KM}	}&{	0.4103 			}&{	0.6103 			}&{	0.3826 			}&{	0.8474 			}&{	0.3244 			}&{	0.5771 			}&{	0.4229 			}&{	0.8369 			}&{	0.6907 			}&{	0.7632 			}&{	\textcolor[rgb]{0.00,0.00,0.63}{0.5866} 	}\\								
\textsf{	\textcolor[rgb]{0.00,0.25,0.50}{CDL+KM	}}&{	0.4138 			}&{	0.5632 			}&{	0.4217 			}&{	0.6490 			}&{	0.2132 			}&{	0.5547 			}&{	0.6893 			}&{	0.9023 			}&{	0.5858 			}&{	0.8293 			}&{	\textcolor[rgb]{0.00,0.00,0.63}{0.5822 }	}\\								
\textsf{	\textcolor[rgb]{0.00,0.25,0.50}{VGAE+KM}	}&{	0.4695 			}&{	0.5710 			}&{	0.3975 			}&{	0.6538 			}&{	0.3307 			}&{	0.6085 			}&{	0.5503 			}&{	0.9023 			}&{	0.6549 			}&{	0.7264 			}&{	\textcolor[rgb]{0.00,0.00,0.63}{0.5865} 	}\\								
\hline\textsf{	\textbf{\textcolor[rgb]{0.00,0.25,0.50}{MC-AE+KM}}	}&{	0.4010 			}&{	0.5615 			}&{	0.3798 			}&{	0.6527 			}&{	\textbf{0.5249 }			}&{	0.6038 			}&{	\textbf{0.9853 	}		}&{	\textbf{0.9390 }			}&{	0.5839 			}&{	\textbf{0.8658 	}		}&{	\textbf{\textcolor[rgb]{0.00,0.00,0.63}{0.6498} }	}\\								
\hline\textsf{	\textcolor[rgb]{0.00,0.25,0.50}{DAE+SC}	}&{	0.6533 			}&{	0.7431 			}&{	0.5977 			}&{	0.6626 			}&{	0.4452 			}&{	0.7699 			}&{	0.7308 			}&{	0.6864 			}&{	\textbf{0.7938 }			}&{	0.7844 			}&{	\textcolor[rgb]{0.00,0.00,0.63}{0.6867 }	}\\								
\textsf{	\textcolor[rgb]{0.00,0.25,0.50}{Deep-FS+SC}	}&{	0.6323 			}&{	0.5429 			}&{	0.6366 			}&{	0.9066 			}&{	0.3675 			}&{	0.7462 			}&{	0.6769 			}&{	0.8467 			}&{	0.7761 			}&{	0.8312 			}&{	\textcolor[rgb]{0.00,0.00,0.63}{0.6963} 	}\\								
\textsf{	\textcolor[rgb]{0.00,0.25,0.50}{fGraphDBN+SC}	}&{	0.6500 			}&{	0.6479 			}&{	0.6295 			}&{	\textbf{0.9156 	}		}&{	0.4068 			}&{	0.7075 			}&{	0.4742 			}&{	0.7011 			}&{	0.6907 			}&{	0.7633 			}&{	\textcolor[rgb]{0.00,0.00,0.63}{0.6587 }	}\\								
\textsf{	\textcolor[rgb]{0.00,0.25,0.50}{CDL+SC}	}&{	0.6183 			}&{	0.7079 			}&{	0.6995 			}&{	0.6143 			}&{	0.3981 			}&{	0.7349 			}&{	0.7101 			}&{	0.9002 			}&{	0.7588 			}&{	0.6268 			}&{	\textcolor[rgb]{0.00,0.00,0.63}{0.6769} 	}\\								
\textsf{	\textcolor[rgb]{0.00,0.25,0.50}{VGAE+SC	}}&{	0.4851 			}&{	0.5894 			}&{	0.4343 			}&{	0.7360 			}&{	0.3926 			}&{	0.6254 			}&{	0.5601 			}&{	0.9236 			}&{	0.7169 			}&{	0.8213 			}&{	\textcolor[rgb]{0.00,0.00,0.63}{0.6285 }	}\\								
\hline\textsf{	\textbf{\textcolor[rgb]{0.00,0.25,0.50}{MC-AE+SC}}	}&{	\textbf{0.6533 }			}&{\textbf{	0.7429 	}		}&{	\textbf{0.7346 }			}&{	0.9140 			}&{	0.4398 			}&{	0.7699 			}&{	0.7282 			}&{	0.8612 			}&{	\textbf{0.7938 	}		}&{	0.6771 			}&{	\textbf{\textcolor[rgb]{0.00,0.00,0.63}{0.7315} }	}\\																						
\hline
\end{tabular}}
\end{center}

\end{sidewaystable}

\begin{table*}[]
\scriptsize
 \centering
 \caption{results of the K-means and SC algorithms on ten real-valued datasets.}
 \label{Tab03}
 \begin{tabular}{cccccccccc}

 \toprule

 \multirow{4}{*}{\textbf{Dataset}} & \multicolumn{2}{c}{\textbf{$\textcolor[rgb]{0.00,0.00,0.63}{\overline{ACC}}$}} & \multicolumn{2}{c}{\textbf{$\textcolor[rgb]{0.00,0.00,0.63}{\overline{Jac}}$}} & \multicolumn{2}{c}{\textbf{$\textcolor[rgb]{0.00,0.00,0.63}{\overline{FMI}}$}}  \\

 \cmidrule(r){2-3} \cmidrule(r){4-5} \cmidrule(r){6-7}

 &   K-means  &   SC &  K-means  &   SC & K-means  &   SC  \\
 \midrule
 \textsf{banner}  & {0.4667$\pm$0.0192}& {0.3713$\pm$0.0158}& {0.3535}& {0.3251} & {0.5702} & {0.5470} \\
 \textsf{beret}  & {0.4482$\pm$0.0111}& {0.3893$\pm$0.0060}& {0.3031}& {0.2631}& {0.4746} & {0.4282} \\
 \textsf{bugatti}  & {0.4240$\pm$0.0346}& {0.3813$\pm$0.0114}& {0.2760} & {0.2643} & {0.4423} & {0.4302} \\
 \textsf{building}   & {0.5799$\pm$0.0026}& {0.4724$\pm$0.0023}& {0.3582}& {0.2917} & {0.5385} & {0.4661} \\
\textsf{vista}  & {0.4698$\pm$0.0026}& {0.4693$\pm$0.0185}& {0.2840}& {0.2802}  & {0.4479} & {0.4436} \\
 \textsf{vistawallpaper} & {0.4781$\pm$0.0118} & {0.4610$\pm$0.0158}& {0.2896} & {0.2807}  &{0.4545} & {0.4442} \\
 \textsf{voituretuning}   & {0.4407$\pm$0.0112}& {0.3561$\pm$0.0039}& {0.2867}& {0.2444} &   {0.4469}&   {0.3990} \\
 \textsf{water}   & {0.4071$\pm$0.0014}&  {0.4255$\pm$0.0006}& {0.2385}& {0.2411} &   {0.3883}  & {0.3920} \\
 \textsf{wing}   & {0.4677$\pm$0.0039}& {0.3929$\pm$0.0208}& {0.2800} & {0.2481} &   {0.4398} & {0.4035}\\
 \textsf{worldmap}  & {0.4595$\pm$0.0127}& {0.3872$\pm$0.0047}& {0.3014}& {0.2760}  &   {0.4710}  & {0.4456} \\

\hline  {\textbf{Average}} & {0.4642}& {0.4106} & {0.2971}&{0.2715} &  {0.4674} & {0.4399}\\

 \bottomrule

 \end{tabular}
\end{table*}

\begin{table*}[]
\scriptsize
 \centering
 \caption{The results of the K-means and SC algorithms on ten UCI datasets.}
 \label{Tab03}
 \begin{tabular}{cccccccccc}
 \toprule

\multirow{4}{*}{\textbf{Dataset}} & \multicolumn{2}{c}{\textbf{$\textcolor[rgb]{0.00,0.00,0.63}{\overline{ACC}}$}} & \multicolumn{2}{c}{\textbf{$\textcolor[rgb]{0.00,0.00,0.63}{\overline{Jac}}$}} & \multicolumn{2}{c}{\textbf{$\textcolor[rgb]{0.00,0.00,0.63}{\overline{FMI}}$}}  \\

 \cmidrule(r){2-3} \cmidrule(r){4-5} \cmidrule(r){6-7}

 &   K-means  &   SC &  K-means  &   SC & K-means  &   SC  \\

 \midrule
 \textsf{balance}  & {0.5221$\pm$0.0544} & {0.5173$\pm$0.0406 } & {0.3747}& {0.2727} & {0.5492}  & {0.4321} \\
 \textsf{biodegradation}  & {0.5886$\pm$0.0000} & {0.6253$\pm$0.0005  } & {0.4179}& {0.5174}& {0.5906} & {0.7046}  \\
  \textsf{car} & {0.4282$\pm$0.0412}  & {0.4088$\pm$0.0724 } & {0.2864}& {0.2716} & {0.4736}  & {0.4573} \\
 \textsf{ClimateModel} & {0.5383$\pm$0.0312} & {0.7827$\pm$0.0011 } & {0.4578}& {0.6464} & {0.6495} & {0.7869} \\
  \textsf{dermatology}  & { 0.2942$\pm$0.0274 }  & {0.3087$\pm$0.0047} & {0.1488} & {0.1370}& {0.2594} & {0.2412} \\
\textsf{HabermanSurvival} & {0.5131$\pm$0.0113 }  & {0.5218$\pm$0.0019} & {0.3776}& {0.3787} &{0.5509}  & {0.5520}  \\
  \textsf{Kdd}  & { 0.6622$\pm$ 0.0005}  & {0.6845$\pm$0.0004} & {0.5644}& {0.5725} &   {0.7490} & {0.7684}  \\
 \textsf{OLD}& {0.9017$\pm$0.0000}  & {0.9056$\pm$0.0001} & {0.8215}& {0.8253}  & {0.9023} & {0.9031} \\
\textsf{parkinsons} & {0.5436$\pm$0.0000}  & {0.5795$\pm$0.0000} & {0.4152} & {0.4134} &   {0.5872}  & {0.5861}  \\
 \textsf{secom} & {0.7571$\pm$0.0004}   & { 0.5110$\pm$0.0075} & {0.6157} & {0.4667}&   {0.7688} & {0.6616}  \\
\hline  {\textbf{Average}} & { 0.5749} & {0.5845} & {0.4480}& { 0.4502 } &  { 0.6080}  & {0.6111} \\
\bottomrule

 \end{tabular}
\end{table*}

\begin{table*}[]
\scriptsize
 \centering
 \caption{The CPU times (second) of the MC-AE+KM and MC-AE+SC algorithms on ten real-valued datasets. The CPU times of MC-AE+KM algorithm consists of the training time of MC-AE model and clustering time of K-means algorithm. The CPU times of MC-AE+SC algorithm consists of the training time of MC-AE model and clustering time of SC algorithm.}
 \label{Tab03}
 \begin{tabular}{cccccccccc}
 \toprule

{\textbf{Dataset}} &   K-means  &   SC & MC-AE  & MC-AE+KM  &   MC-AE+SC  \\

 \midrule
 \textsf{banner}  & {	2.4688 	}  & {	1.5938 	}  & {	7216.2500 	}  & {	7218.7188 	}  & {	7217.8438 	}  \\
\textsf{beret}  & {	2.0156 	}  & {	1.8594 	}  & {	7215.3750 	}  & {	7217.3906 	}  & {	7217.2344 	}  \\
\textsf{bugatti}  & {	2.8438 	}  & {	2.2500 	}  & {	7154.9531 	}  & {	7157.7969 	}  & {	7157.2031 	}  \\
\textsf{building}   & {	3.4688 	}  & {	1.4063 	}  & {	7481.1875 	}  & {	7484.6563 	}  & {	7482.5938 	}  \\
\textsf{vista}  & {	2.1875 	}  & {	1.2656 	}  & {	6745.1563 	}  & {	6747.3438 	}  & {	6746.4219 	}  \\
\textsf{vistawallpaper} & {	2.1875 	}  & {	1.4688 	}  & {	6718.3281 	}  & {	6720.5156 	}  & {	6719.7969 	}  \\
\textsf{voituretuning}   & {	2.0625 	}  & {	2.0781 	}  & {	7235.9219 	}  & {	7237.9844 	}  & {	7238.0000 	}  \\
\textsf{water}   & {	3.0313 	}  & {	1.5469 	}  & {	7807.7969 	}  & {	7810.8281 	}  & {	7809.3438 	}  \\
\textsf{wing}   & {	2.3438 	}  & {	1.4844 	}  & {	7052.0000 	}  & {	7054.3438 	}  & {	7053.4844 	}  \\
\textsf{worldmap}  & {	2.1250 	}  & {	1.8750 	}  & {	7804.4063 	}  & {	7806.5313 	}  & {	7806.2813 	}  \\

\bottomrule

 \end{tabular}
\end{table*}

\begin{table*}[]
\scriptsize
 \centering
 \caption{The CPU times (second) of the MC-AE+KM and MC-AE+SC algorithms on ten UCI datasets. The CPU times of MC-AE+KM algorithm consists of the training time of MC-AE model and clustering time of K-means algorithm. The CPU times of MC-AE+SC algorithm consists of the training time of MC-AE model and clustering time of SC algorithm.}
 \label{Tab03}
 \begin{tabular}{cccccccccc}
 \toprule

{\textbf{Dataset}} &   K-means  &   SC & MC-AE  & MC-AE+KM  &   MC-AE+SC  \\

 \midrule
 \textsf{balance}  & {	0.0469 	}  & {	0.5156 	}  & {	6.1719 	}  & {	6.2188 	}  & {	6.6875 	}  \\
\textsf{biodegradation}  & {	0.3594 	}  & {	0.5313 	}  & {	47.7500 	}  & {	48.1094 	}  & {	48.2813 	}  \\
  \textsf{car} & {	0.3125 	}  & {	2.0156 	}  & {	17.7969 	}  & {	18.1094 	}  & {	19.8125 	}  \\
 \textsf{ClimateModel} & {	0.3750 	}  & {	0.8438 	}  & {	11.0625 	}  & {	11.4375 	}  & {	11.9063 	}  \\
\textsf{dermatology}  & {	0.2344 	}  & {	0.6094 	}  & {	12.4844 	}  & {	12.7188 	}  & {	13.0938 	}  \\
\textsf{HabermanSurvival} & {	0.0321 	}  & {	0.3594 	}  & {	7.6250 	}  & {	7.6571 	}  & {	7.9844 	}  \\
\textsf{Kdd}  & { 	0.0625 	}  & {	0.0469 	}  & {	2.4688 	}  & {	2.5313 	}  & {	2.5156 	}  \\
\textsf{OLD}& {	1.2031 	}  & {	4.6875 	}  & {	213.8750 	}  & {	215.0781 	}  & {	218.5625 	}  \\
\textsf{parkinsons} & {	0.6094 	}  & {	0.7813 	}  & {	51.3750 	}  & {	51.9844 	}  & {	52.1563 	}  \\
\textsf{secom} & {	2.6406 	}  & {	2.8438 	}  & {	6417.4219 	}  & {	6420.0625 	}  & {	6420.2656 	}  \\

\bottomrule

 \end{tabular}
\end{table*}

\section*{References}

\bibliography{rbm}

\end{document}